\documentclass[letterpaper, 10 pt, conference]{ieeeconf}  
\usepackage{amsmath}
\usepackage{graphicx}
\usepackage{url}
\usepackage{mars}
\IEEEoverridecommandlockouts                              

\usepackage{subfig}
\title{\LARGE \bf
Incrementally Building Topology Graphs via Distance Maps
}

\author{Yijun Yuan$^{1}$ and S\"oren Schwertfeger$^{1}$
\thanks{$^{1}$Both authors are with the School of Information Science and Technology, 
ShanghaiTech University, China.
        {\tt\small [yuanwj, soerensch]@shanghaitech.edu.cn}}%
}

\begin{document}
%
%


\marsPublishedIn{Accepted for:} 		

\marsVenue{IEEE International Conference on Real-time Computing and Robotics(RCAR) 2019}

\marsYear{2019}

\marsPlainAutors{Yijun Yuan and S\"oren Schwertfeger}


\marsMakeCitation{Incrementally Building Topology Graphs via Distance Maps}{IEEE Press}


\marsIEEE{}


\makeMARStitle

%
%
\setlength{\belowcaptionskip}{-5pt}

\maketitle
\thispagestyle{empty}
\pagestyle{empty}

\begin{abstract}
Mapping is an essential task for mobile robots and topological representation often works as a basis for the various applications. In this paper, a novel framework that can build topological maps incrementally is proposed. The algorithm is using a distance map, and in our framework the topological map can grow as we append more sensor data to the map. To demonstrate our algorithm, we show the result of the distance map based method on several popular maps and run the incremental framework with raw sensor data to have a growing topological map, as an example of a robot exploring the environment.
\end{abstract}


\section{INTRODUCTION}
Occupancy Grid Maps and Topological Maps are two major representations for 2D mapping. Certainly, the grid map can provide more specific descriptions of the surroundings, but when it comes to the large-scale environments, the computation cost of applications, planning as an example, will make grid maps inefficient. Topological maps are much smaller and thus faster to compute on representations. 

Topological map generation algorithms that build on top of the Voronoi diagram are mostly widely seen.

From our survey, the approaches of topology map construction can mainly be classified into three categories: 
(1) Use Voronoi diagram vertexes as nodes of topology graph. \cite{setalaphruk2003robot} keeps the landmark such as intersection, corners, and dead-end in corridors as a node, and the Voronoi path between adjacent nodes as edges, to represent a topological map. To make it more presentable, it appends more properties like cross shape intersection into nodes. \cite{wallgrun2004hierarchical} proposes a method to derive a route graph from the generalized Voronoi diagram. A route graph is a special kind of topology map with the vertex as navigational decision points and edges as route segment to those points. 
This algorithm is able to incrementally construct the route graph. 
(2) Split map on the gateway of Voronoi points that is too close to the obstacle. \cite{thrun1998learning} concentrates on the large-scale environment. Here a trade-off will exist between representation efficiency and cost with grid-based maps and topological maps. The author proposes an approach to integrating grid-based maps and topological maps.  By cutting on the gateway of a grid map from the Voronoi diagram, the space can be partitioned into disjoint regions. Then, by mapping them into an isomorphic graph, each vertex will be able to denote one region. \cite{wurm2008coordinated} aims to do robot exploration with segmentation of the environment. Also, with the partition on the critical points, space can be separated into regions. This graph also uses regions as nodes. 
(3) Voronoi random fields (VRF): Integrating the features from grid maps and Voronoi graphs, \cite{friedman2007voronoi} converts the Voronoi graph into a conditional random field (CRF). They solve CRF to segment the environment into regions that might be rooms, junctions, and doorways. Instead of manually tuning the parameters, they learn the model from human-labeled training data. 
Cell Decomposition \cite{acar2002sensor} is another way to generate topological structures. But it is very specific to the task of coverage path planning and does segment the space against the human intuition and is also not rotation independent.

\begin{figure}[tpb]
	\centering
	\subfloat[Grid map]{
		\label{fig:origin}
		\includegraphics[width=0.33\linewidth]{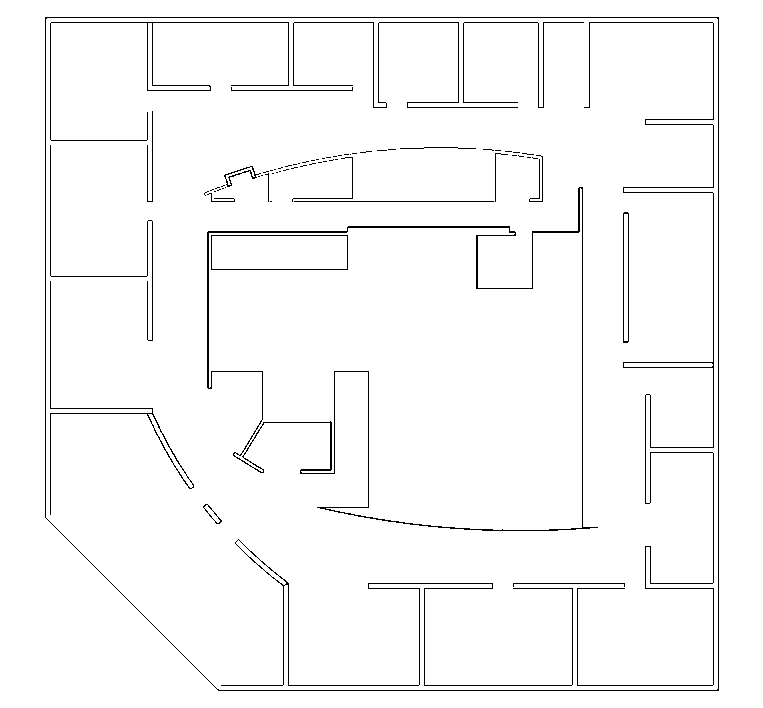}}
	\subfloat[Distance Map]{
		\label{fig:gs}
		\includegraphics[width=0.33\linewidth]{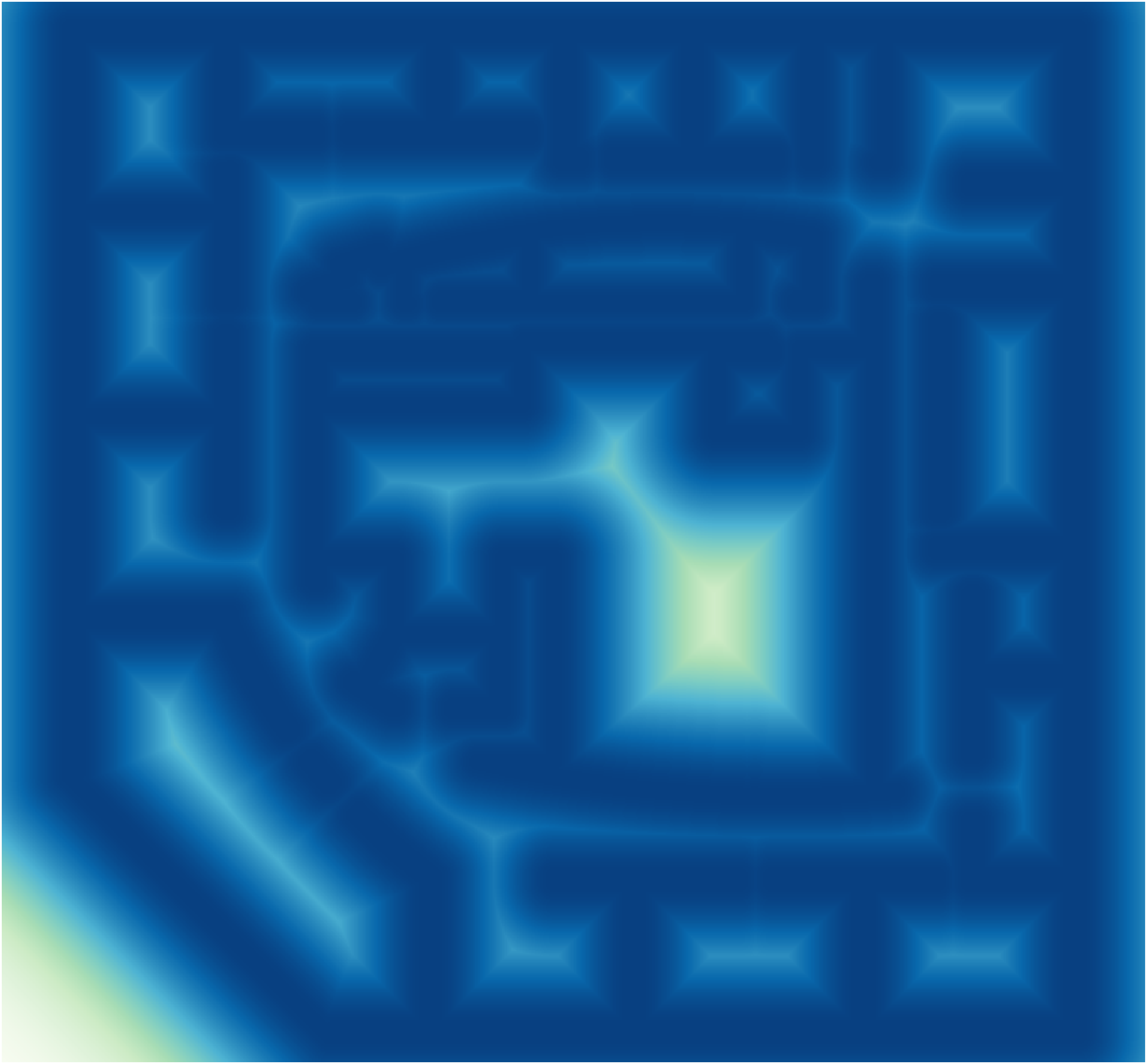}}
	\subfloat[After edge process]{
		\label{fig:lap}
		\includegraphics[width=0.33\linewidth]{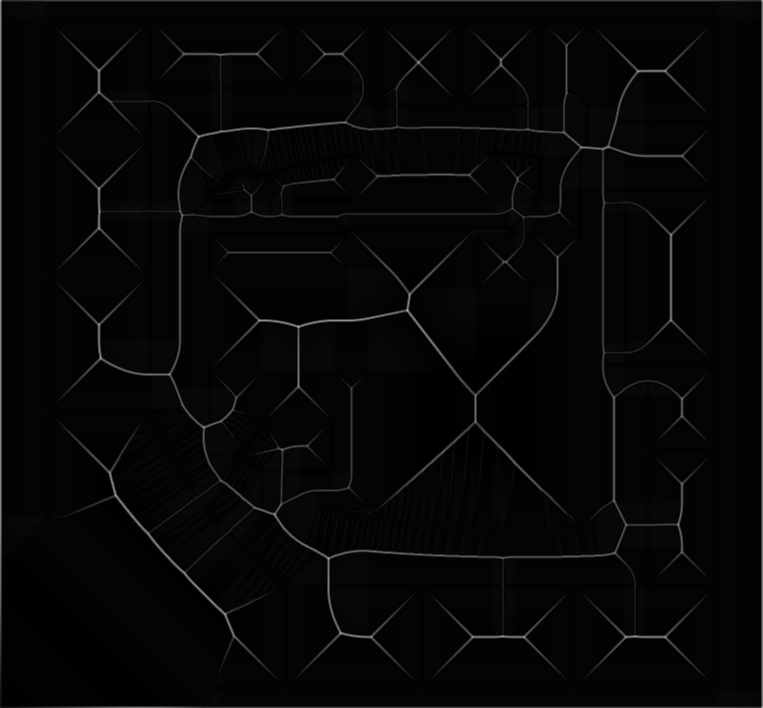}}
	\\
	\subfloat[Binary map]{
		\label{fig:bin_lap}
		\includegraphics[width=0.33\linewidth]{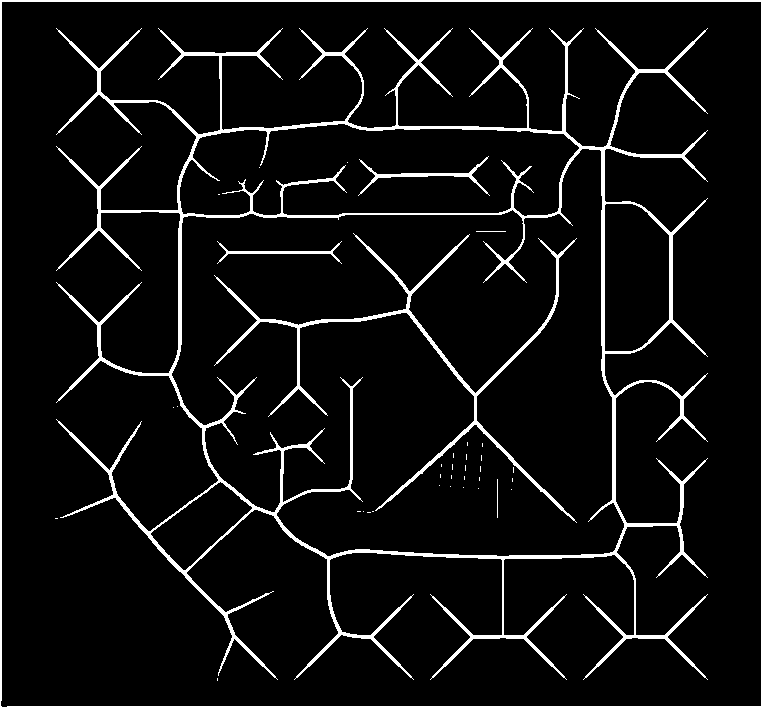}}
	\subfloat[Skeleton]{
		\label{fig:skel}
		\includegraphics[width=0.33\linewidth]{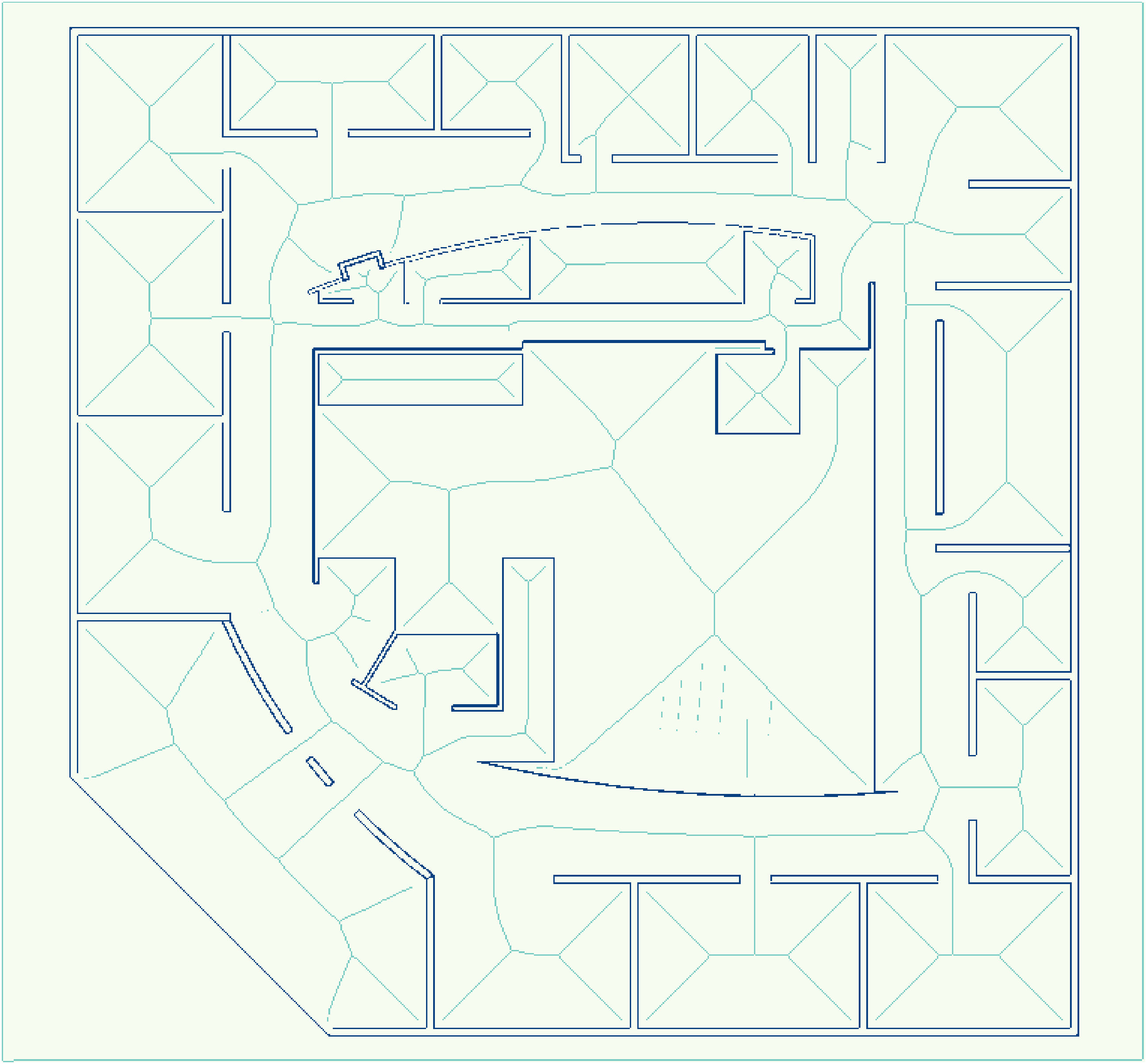}}
	\subfloat[Topological map]{
		\label{fig:topology}
		\includegraphics[width=0.33\linewidth]{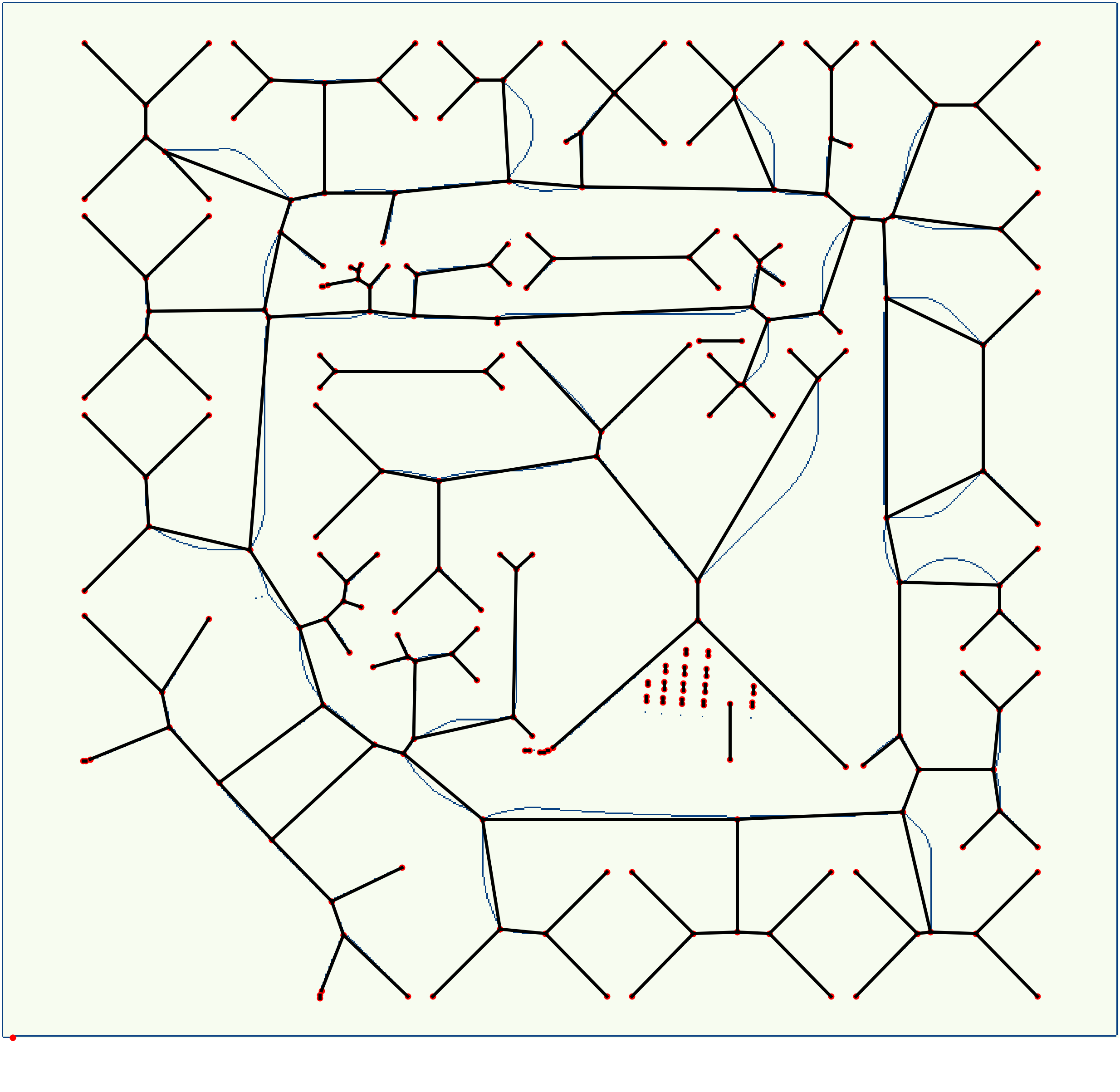}}
	\caption{Topological map generation via Distance map. Fig. \ref{fig:origin} is the input grid map. Fig. \ref{fig:gs}-\ref{fig:skel} are sequentially produced as in Section. \ref{sec:2}. Fig. \ref{fig:topology} is the result topological map.}
	\label{fig:toporep}
\end{figure}

Inspired by the paper \cite{diosi2005interactive}, that utilizes distance transform on grid maps to further employ room segmentation, we consider it should also be possible to extract topological maps from the distance map, because there already exist methods to extract skeletons from distance maps, such as \cite{chang2007extracting}. The skeleton then represents the paths we need to generate a Topology Graph.

In this paper, we first generate the topological map from the distance map. Then we move one step further to explore the incremental topological map building from raw sensor data. Simultaneous Localization and Mapping (SLAM) is widely used in robotics today. 
This paper deals with the mapping part of SLAM and assumes that the localization part is solved within the SLAM framework already.

Unlike \cite{schwertfeger2016map, schwertfeger2013evaluation}, that build topological maps using the Voronoi diagram algorithm, we use the distance map to achieve the skeleton that has a similar shape as the figures in \cite{schwertfeger2016map}. We then use the skeleton to extract the topological map.

 The important process for the above task is to extract a skeleton that is 1 pixel wide. In the study of skeletonization, much brilliant works has been proposed. \cite{van2014scikit} provide good examples of how skeletonizing works.  Iteratively removing pixels on the object border, \cite{zhang1984fast} obtained the skeleton when no more pixel can be removed. \cite{lee1994building} collects candidates to be removed and has a double check to preserve the connectivity in each iteration. The medial axis method is on top of distance transform to provide the ridges as a skeleton. \cite{guo1989parallel} is the method we utilized in our framework, which holds the good property that it can preserve the endpoints of medial curves while avoiding the deletion of pixels in the middle of medial curves. This is a property that we want for our algorithm.
 
\begin{figure}[tpb]
	\centering
	\includegraphics[height=1.2\linewidth]{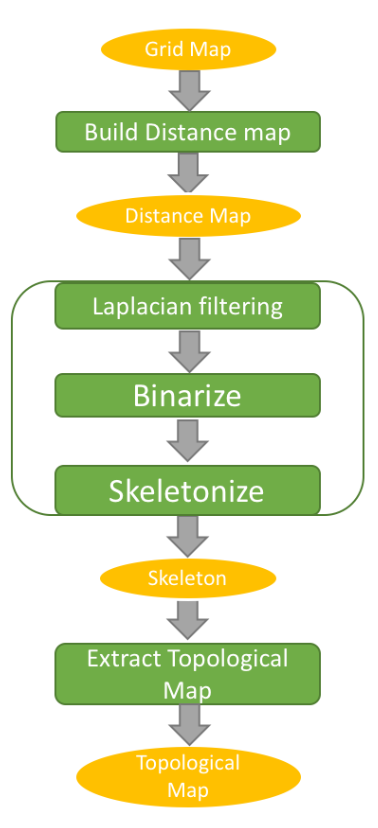}
	\caption{The pipeline of the distance map based method we implemented to generate a topology graph from a grid map. The yellow ellipse is the data while the green window is the process.}
	\label{fig:pipe}
\end{figure}

From the distance map, it should be possible to obtain the skeleton directly with the medial axis method. However, in our incremental skeleton updating, a small layer of skeleton pixel around the new frame should be able to help keep the connectivity. Thus, we prefer to extracting the skeleton from a binary image like reported in \cite{guo1989parallel}, which does meet our needs.

In our framework, all of the distance map, skeleton and topological map will be incrementally updated. 

Actually, \cite{allen2017incremental} has proposed a method that allows for an incremental Voronoi diagram algorithm. However, when it comes to updating the old map with a new scan, Voronoi based methods become complicated, because the obstacle point around in the old map should also be considered. This is expensive, especially for grid maps with too many obstacle points. In contrast, our distance map based method does not have this problem. Using distance maps to represent the spatial information of obstacles, there is no need to retrieve the old observation: We just combine the distance map from the new scan with the old global distance map. 

Though not using the incremental Voronoi diagram algorithm, \cite{van2000incremental} proposed one incremental framework based on Voronoi diagrams. With a similar strategy to update global topological map, \cite{kwon2008real} proposed a framework based on a thinning algorithm. Those two methods keep vertexes in the global map and in the local map, in order to merge the corresponding vertexes upon update.


Our incremental framework has two advantages. First, the position of vertexes in the room region are almost the same as we were directly using the global occupancy map. Additionally, we can achieve high update frequencies, because the frequency does not affect the computation cost of distance map updating in each iteration. 

In our framework, the skeleton is from a distance map, and the topological map is from the skeleton, which means, each part won't be affected by its following steps. Thus, it is convenient to set different updating rates for those three. For example, we can update the distance map with every scan (frame), the skeleton every 100 frames, and the topological map every 200 frames. We consider this reasonable, because it is not necessary to renew the topological map every frame. We could grow the topological map in each frame, but generally, we expect the topological map will not be used with a very high frequency, while we still want to represent all sensor data in the map.

In the following we first describe the distance map based method we implemented in Section \ref{sec:2}. After that, the incremental framework will be introduced in Section \ref{sec:incremental}. Then, Section \ref{sec:experiments} shows experiments to demonstrate our work. In the end, we conclude this paper in Section \ref{sec:conclusions} and emphasize our contribution.

%
%
%
%

\section{Topology Graph Generation Using a Distance Map}
\label{sec:2}

The pipeline of the graph generation can be found in Fig. \ref{fig:pipe}. It can be simply divided into three steps: building the distance map, extracting the skeleton and generating the topological map.

\begin{figure*}[tpb]
	\centering
	\includegraphics[width=1.0\linewidth]{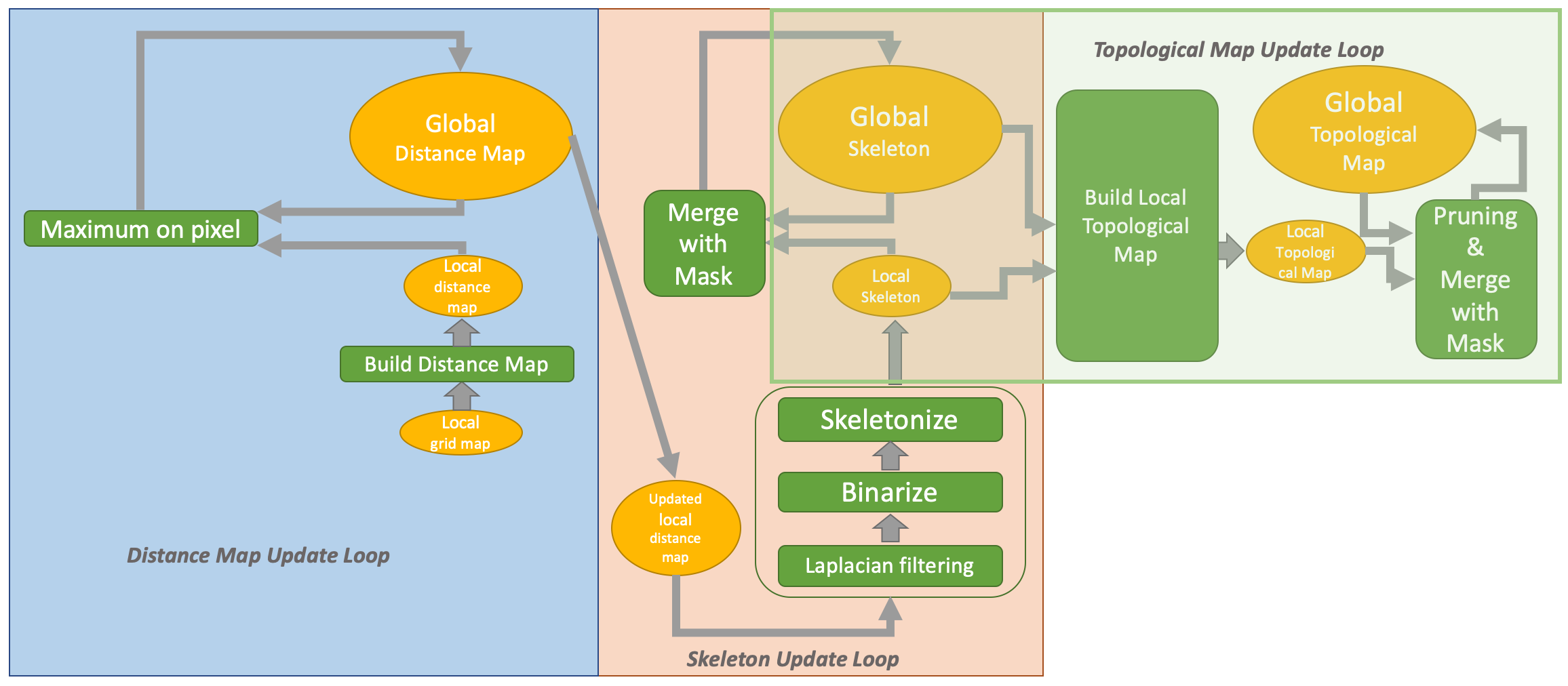}
	\caption{Pipeline of the incremental framework to build a topology graph from raw sensor data. The blue transparent window is for the distance map update loop, the red window is for the skeleton update loop and the green window is for the Topological Map update loop.}
	\label{fig:loop}
\end{figure*}

\subsection{Distance Map}
\label{sec:21}
Firstly we describe how to convert grid occupancy map to a distance map. Generally, the distance map is generated by taking the Euclidean distance to its closest obstacle as its pixel value. The chosen distance metric to obtain the distance map does not have to be Euclidean distance. It could be Manhattan distance, for example. In our work we choose to utilize the Gaussian function by putting the Gaussian kernel on obstacle points to achieve the distance map.

For an occupied pixel on position $p_i$, with $i\in \{1,2,3\cdots N\}$, where $N$ is the number of occupied points, we represent it as a delta function $\delta(x-p_i)$. The function to generate a distance map of each point $i$ as a convolution of $\delta(x-p_i)$ and a Gaussian kernel $G_{\sigma}$ is:
\begin{equation}
H_i(x) = \delta(x-p_i) \ast G_{\sigma}(x).
\end{equation}
Then we compute the maximum as the final distance map value. The final distance map function is:
\begin{equation}
H(x) = \max_{i=1}^{N} H_i(x)
\end{equation}

Here, pixels on the distance map are affected by the Gaussian kernel utilized on each obstacle points, and will be given the maximum value. Thus, each pixel value is dominated by the nearest obstacle point and the function $H$ is selected to achieve our distance map.

For the distance map we created with Gaussian kernel, the closer the pixel is to the obstacle, the larger the value would be. Intuitively, if we consider the pixel value as the third axis, from Fig. \ref{fig:gs}, the distance map with Gaussian function will make the space around the obstacle look like a "valley". And the "rivulet" between the "valley" is the paths we want to extract.

\subsection{Distance Map To Skeleton}
\label{sec:22}
To extract the "rivulet" from the distance map, some filter to extract edges can be used. Here we select the Laplacian filter as the solution. The Laplacian filter is a very useful filter that can be utilized to find the fast changing area. To make it easy to utilize on images, an approximate discrete convolution kernel is utilized. With the process of filtering, we achieve the "rivulet" that is shown in Fig. \ref{fig:lap}.

However, this is not exactly what we need, since only having the "rivulet" can not directly provide the topological map. So suppression is necessary to help achieve a very thin path. \cite{thin} provides several thinning algorithms to skeletonize the binary image. Before that thinning process, we have to binarize the above "rivulet" with a certain threshold, and obtain Fig. \ref{fig:bin_lap}.

Next, the thinning algorithm is applied on the binarized "rivulet", to generate a skeleton. From Fig. \ref{fig:skel} we can see that the skeleton only has a one-pixel width.

It should be noted that a T shape skeleton in an arbitrary $3\times3$ window will cause a redundant edge in following step. So in our implementation, we also remove the cross pixel of the T shape skeleton in that $3\times3$ window.

\subsection{Building the Topology Graph}
\label{sec:23}
Since the skeleton has been obtained, a topological map can be created by assuming a connection between neighboring pixels.

First, we represent each skeleton pixel as a vertex. The vertex will have an edge to each of its up to 8 occupied neighbors. For each skeleton pixel as the center of a $3\times 3$ window, other skeleton points covered with such a window can be considered as its neighbors. However, the $L$ shape skeleton in $2\times2$ window will cause an additional edge on the diagonal, so we will not add these edges. 

Then, we delete all of the degree 2 vertexes by combining its two edges.

One result graph can be found in Fig. \ref{fig:topology}.

%

\section{Incrementally Building the Topological Map with Sensor Data}
\label{sec:incremental}

In the above section, we can build a topological map using a distance map. This is an excellent basis for incrementally building the topological map. As we make the robot move, the topological map can be created with the sensor data as input. Fig. \ref{fig:loop} illustrates the pipeline for this framework. It consists of three parts: (1) distance map update loop, (2) skeleton update loop and (3) topological map update loop.

To make it easy to demonstrate, we use "local" and "global" to name the map for newly observed data and the preserved map that is growing as robot moves and scans, respectively. 

\subsection{Distance Map Update Loop}
In the distance map update loop, for each frame $t$, with the odometry and laser data, we can compute the locations of the obstacle points. 
Then the local distance map for this frame can be generated with the method described in Section \ref{sec:21}.

By combining the global distance map function $H^{t-1}$ and local distance map function $\hat{H}^t$, we achieve the new global distance map with the updated function
\begin{equation}
H^t(x) = \max( H^{t-1}(x), \hat{H}^t(x-c_t))
\end{equation}
, where $c_t$ is the shift (offset) from the local map coordinate to the global map coordinate.

Actually, during the implementation, we just take the maximum between {\tt global} and {\tt local distance map} in each pixel for new {\tt global distance map}.
Thus the distance map can be obtained incrementally.

\begin{figure}[tpb]
	\centering
	\subfloat[]{
		\label{fig:ine}
		\includegraphics[width=0.42\linewidth]{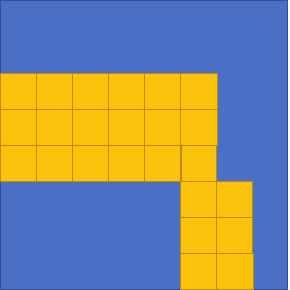}}
	\subfloat[]{
		\label{fig:oute}
		\includegraphics[width=0.42\linewidth]{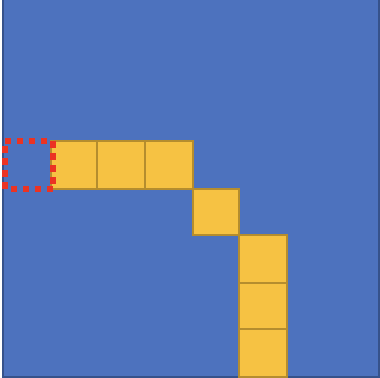}}\\
	\subfloat[]{
		\label{fig:msk}
		\includegraphics[width=0.42\linewidth]{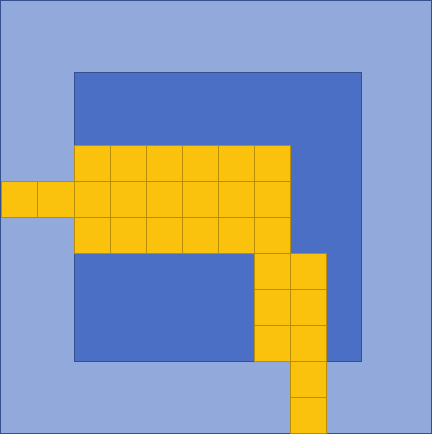}}
	\subfloat[]{
		\label{fig:msk_out}
		\includegraphics[width=0.42\linewidth]{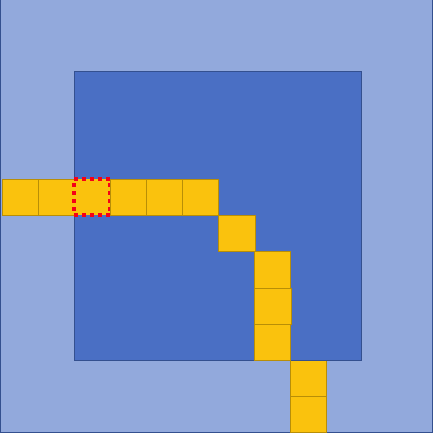}}
	\caption{Keep the connectivity with the skeleton layer. Fig. \ref{fig:ine}, \ref{fig:oute} / Fig. \ref{fig:msk}, \ref{fig:msk_out} are without / with appended outer layer. The problematic pixel is marked with a red, dashed frame. Fig. \ref{fig:ine}, \ref{fig:msk} / Fig. \ref{fig:oute}, \ref{fig:msk_out} are the input binary map / output skeleton. Yellow pixel are binary true. Blue / Light blue space is masked local space / appended layer.}
	\label{fig:thinning}
\end{figure}

\subsection{Skeleton Update Loop}
In our framework, the loop to update the skeleton is shown in the red window of Fig. \ref{fig:loop}. We keep one mask to indicate the area of the global distance map that has been updated since the last skeleton update. 
Using the masked region we obtain the updated parts of the {\tt global distance map}, which then is our {\tt local distance map}. The next step is to extract the local skeleton. But it is not that straightforward to apply the method in Section \ref{sec:22} on the updated {\tt local distance map}. 

From the previous global skeleton, some parts of the skeleton outside of the masked space may stretch into the masked space. So directly applying the algorithm in Section \ref{sec:22} will cause the deletion of pixels in branches which should be preserved.

We combine the {\tt local binary map} with a small layer of the {\tt global skeleton map} around it. Now we need to ensure, that the {\tt local skeleton map} that we create is connected with the old global skeleton layer that we added. To solve this problem, we append a small layer of {\tt global skeleton map}. Then we use the thinning algorithm presented in \cite{guo1989parallel}, that can preserve the endpoints of medial curves while keeping the connectivity of the line. In this step the skeleton will be updated from the {\tt combined binary map} by applying the thinning algorithm on it and renewing the {\tt global skeleton} in the masked space. Comparing Fig. \ref{fig:oute} with Fig. \ref{fig:msk_out}, we can see that the appended layer helps keeping the pixel in red dashed frame to preserve the connectivity.


\subsection{Topological Map Update Loop}
\label{sec:33}

To update the topological map, another mask (local skeleton mask) is utilized to indicate the space that has been updated since the last topological map update. The topological update loop can be found in the green window of Fig. \ref{fig:loop}. It can be divided into five sequential steps at the $t$-th update:
 
1. Build the {\tt local topological graph} $\hat{G}^t$ using the skeleton in the local masked space, by applying the algorithm from Section \ref{sec:23}. Here, the local skeleton mask is the region that the updates of skeleton covered since the last time the {\tt topological map} $G^{t-1}$ was updated. 

2. Trim edges of the {\tt global graph} $G^{t-1}$ that have paths in the local skeleton mask. If an edge is totally covered by the mask, it will be removed. If part of the edge is in the mask, it will be trimmed into several smaller edges that are outside of the mask. So we achieve the {\tt outer mask global graph} $G_{outer}^{t-1}$.

3. Insert edges into the local graph $\hat{G}^t$ that go from the boundary vertexes to the {\tt outer mask global graph} vertexes. In the local graph, keep edges that are not on the boundary of the mask. Then establish the connection between the local graph and the {\tt outer mask global graph} by connecting the edges with boundary vertexes to the {\tt outer mask global graph}  vertexes near the boundary. We call the new {\tt expanded local graph} $\hat{G}_{exp}^t$.

4. Create the union of the {\tt expanded local graph} $\hat{G}_{exp}^t$ and the {\tt outer mask global graph}  $G_{outer}^{t}$ to update the global topological graph. 

5. Post processes to achieve the new {\tt global topology graph} $G^t$ by removing the degree 2 vertexes on the boundary of mask.
\begin{figure}[b!]
	\centering
	\includegraphics[width=0.45\linewidth]{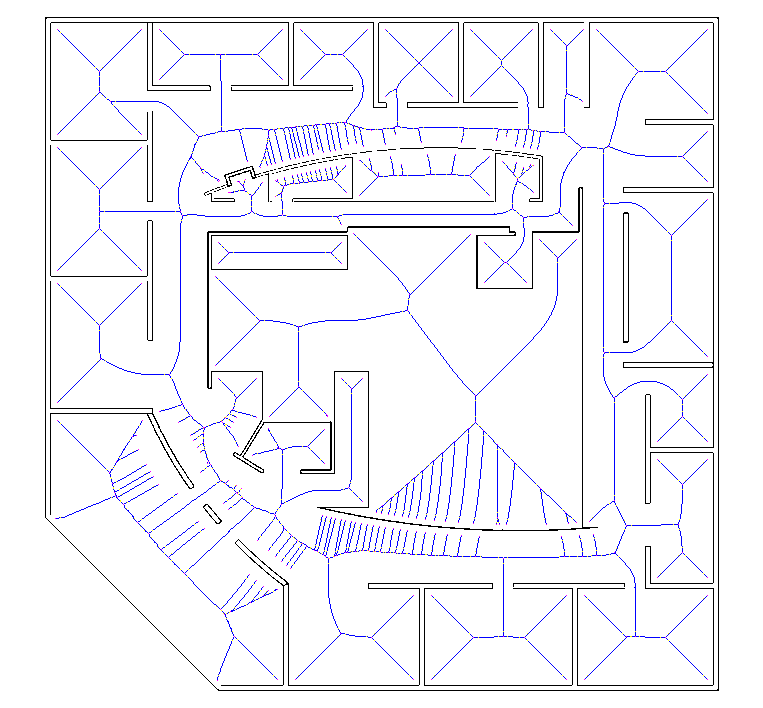}
	\includegraphics[width=0.45\linewidth]{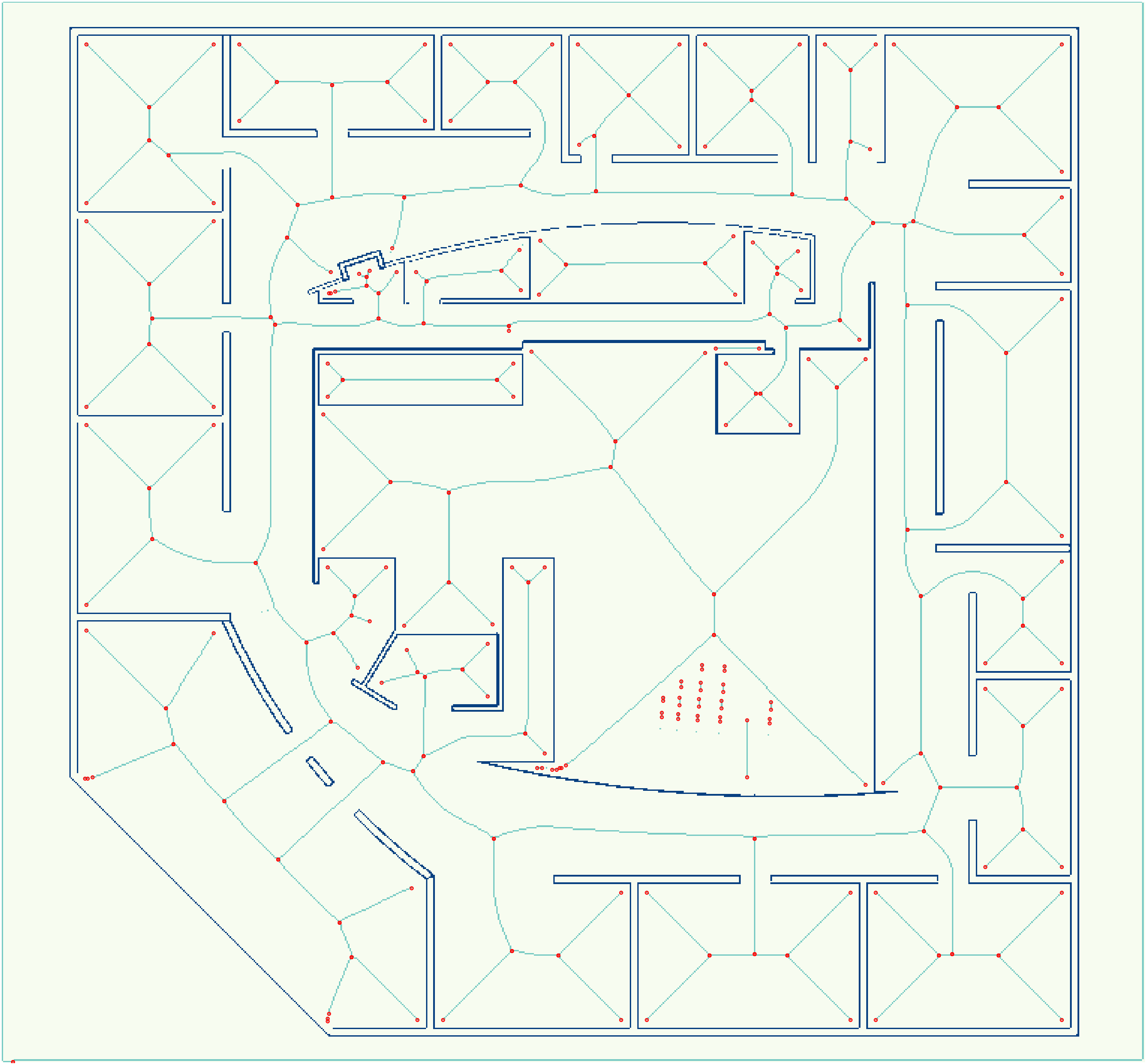}\\ \vspace{3mm}
	\includegraphics[width=0.45\linewidth]{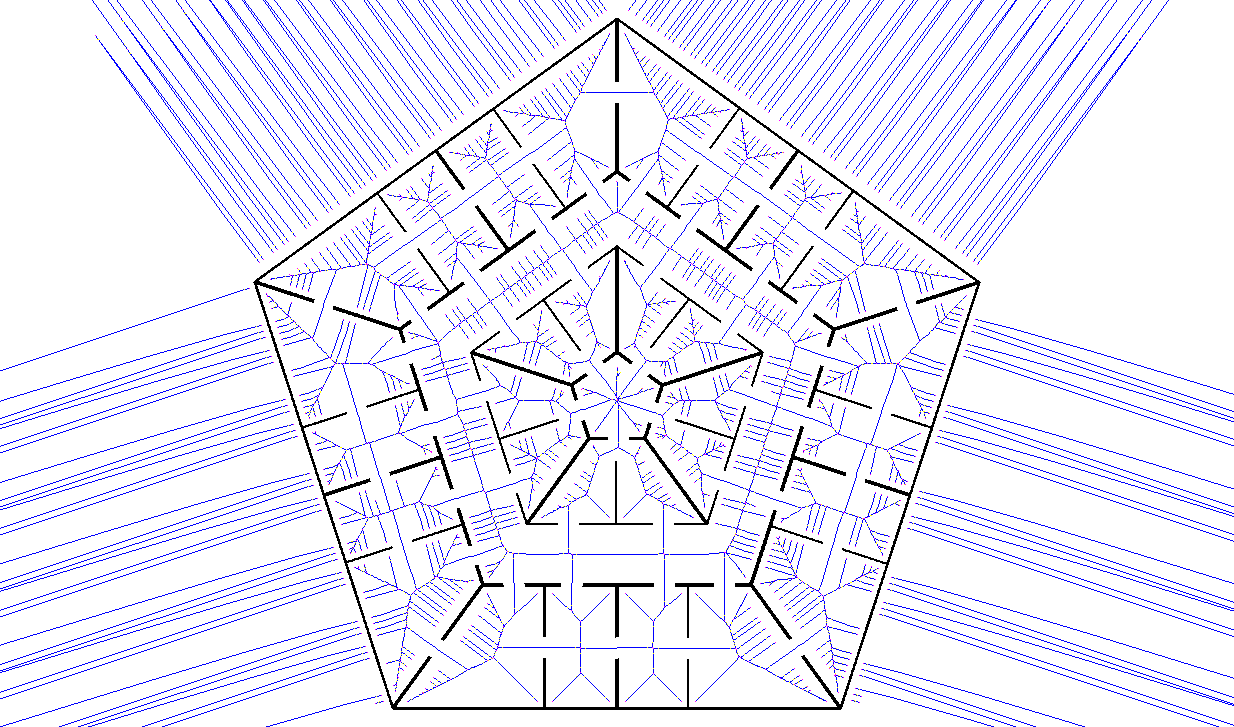}
	\includegraphics[width=0.45\linewidth]{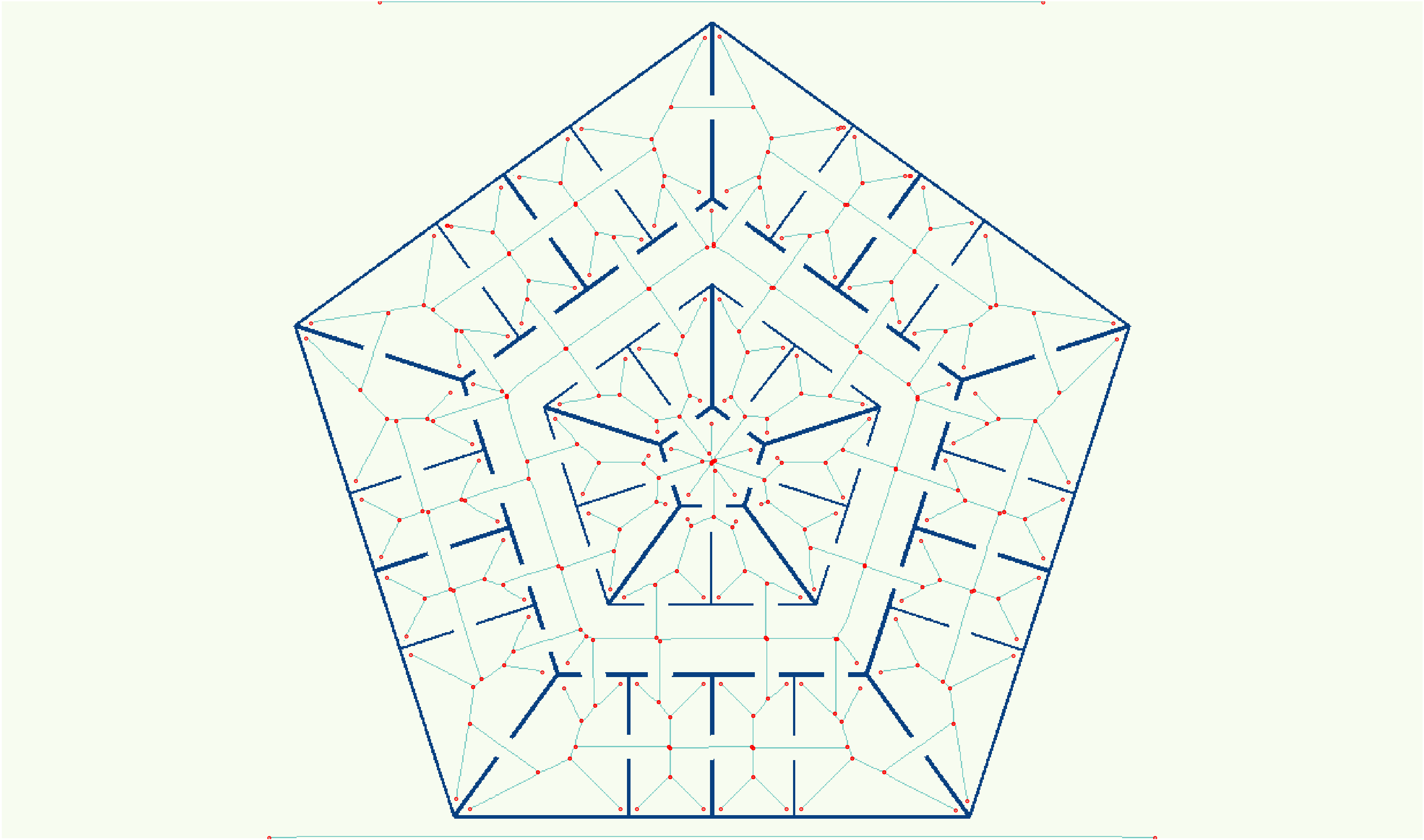}\\\vspace{3mm}
	\includegraphics[width=0.45\linewidth]{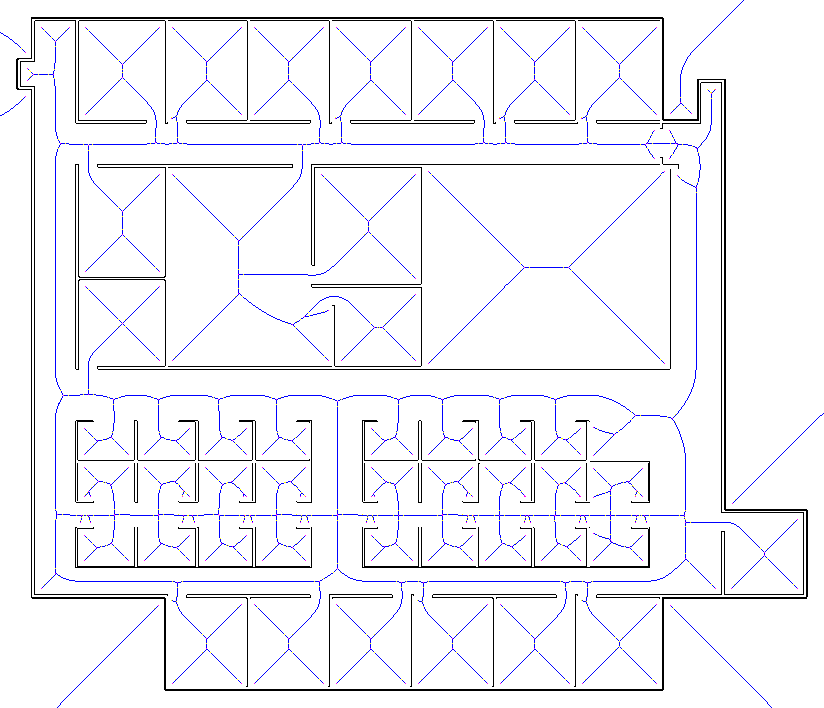}
	\includegraphics[width=0.45\linewidth]{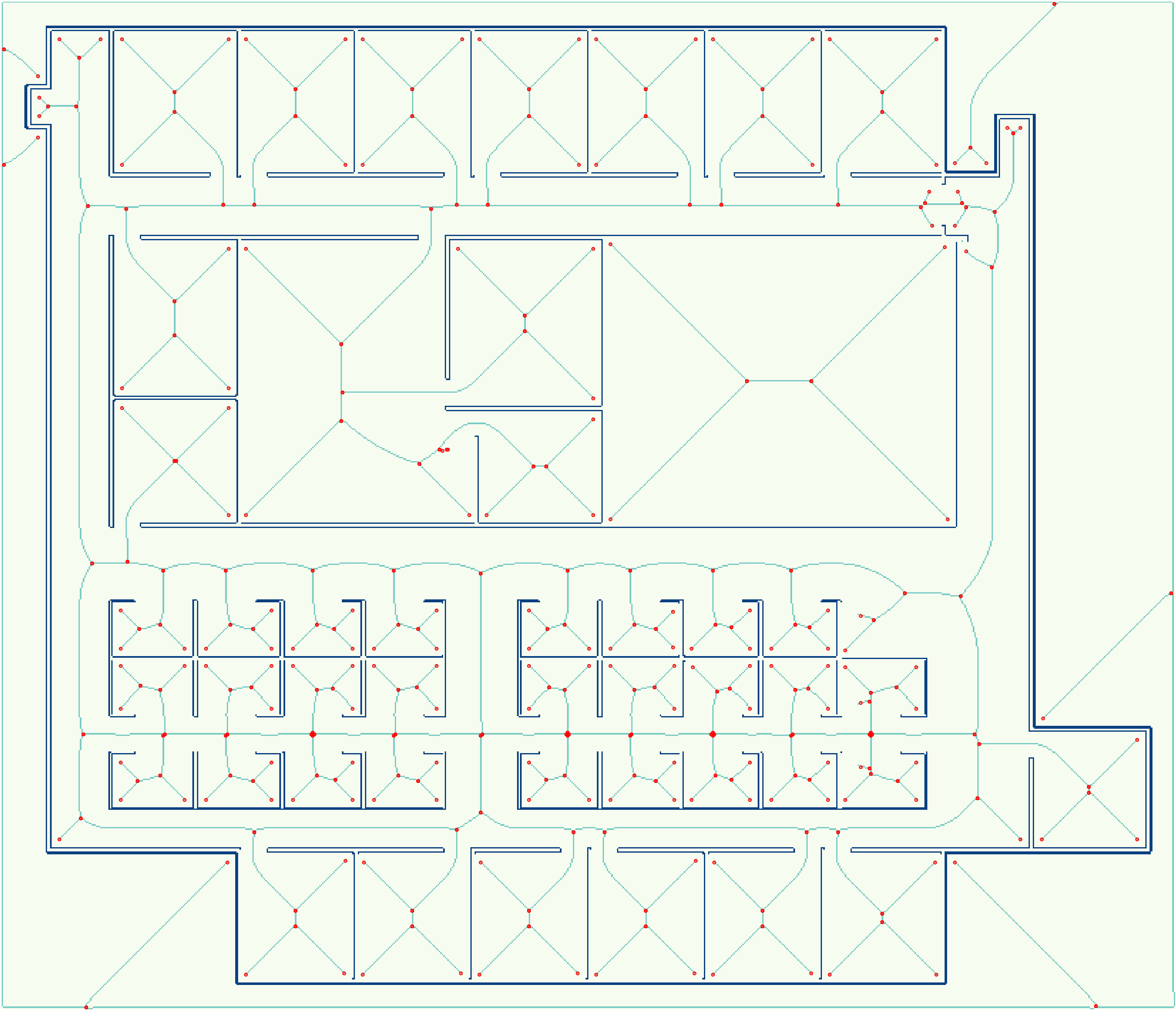}
	\caption{Topological maps annotated with path information. The left column is from the implementation of \cite{schwertfeger2016map} and the right column is the from our implementation of distance map based method.}
	\label{fig:vori_vs_heatmap}
\end{figure}
\begin{table}[b!]
	\caption{Vertex errors for maps in Fig. \ref{fig:vori_vs_heatmap}.}
	\centering
	\begin{tabular}{|l|l|l|l|}
		\hline
		& Ave Dist & Outlier / Total &  \% ($\le$1)\\ \hline
		intel ($763\times708$) &2.17 & 15 / 258 & 72.4\% \\ \hline
		office ($1234\times 727$) &1.35 & 4 / 305 &49.5\% \\ \hline
		a\_scan ($824\times708$) & 0.88 &6 / 344 & 84.9\%\\ \hline
	\end{tabular}
	\label{tab:avg_time}
\end{table}
\section{Experiments}
\label{sec:experiments}

\subsection{Setting}
\label{sec:setting}
\newcommand\blaSize{0.45}
\begin{figure}[b!]
	\centering
	\includegraphics[width=\blaSize\linewidth]{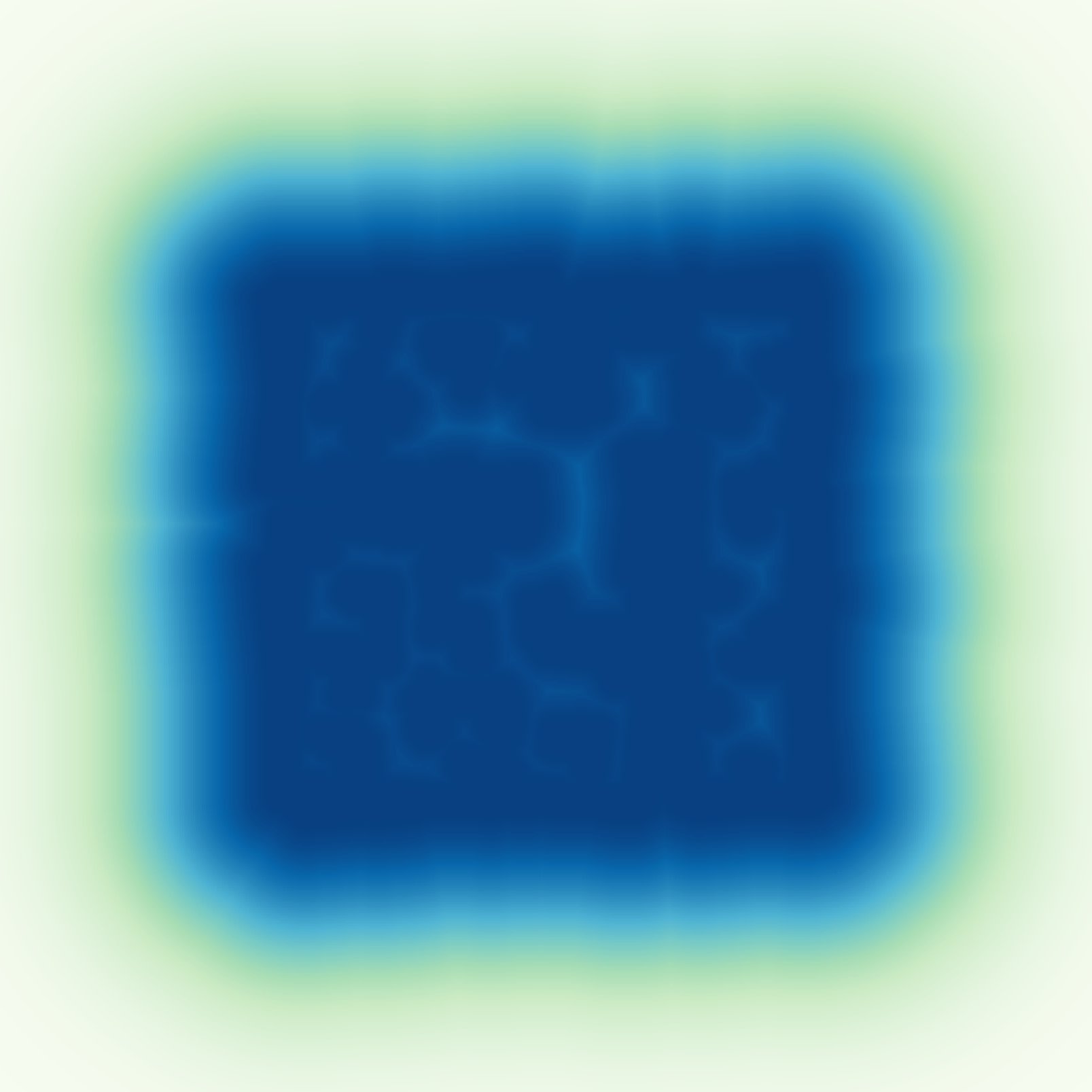}
	\includegraphics[width=\blaSize\linewidth]{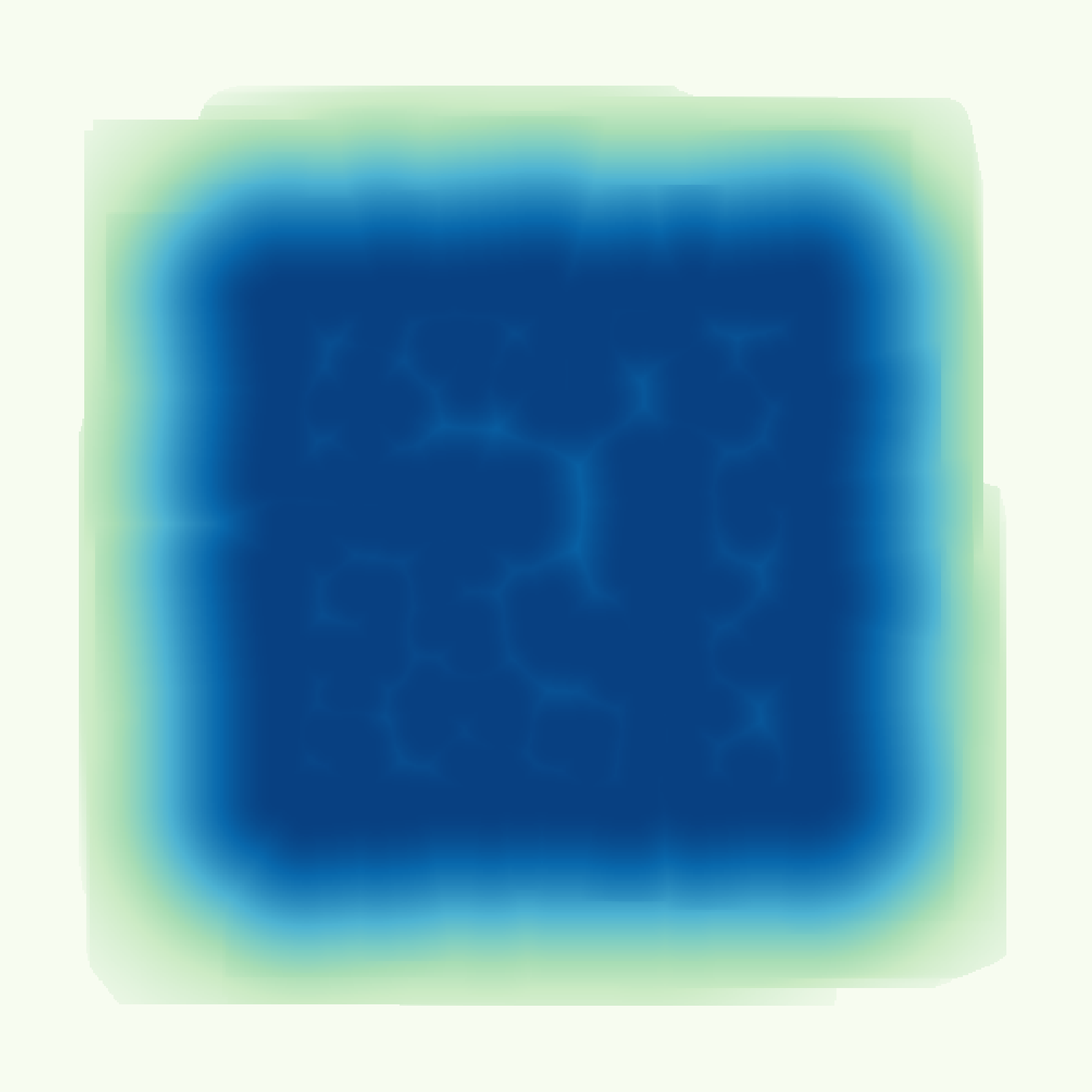}\\ \vspace{1mm}
	\includegraphics[width=\blaSize\linewidth]{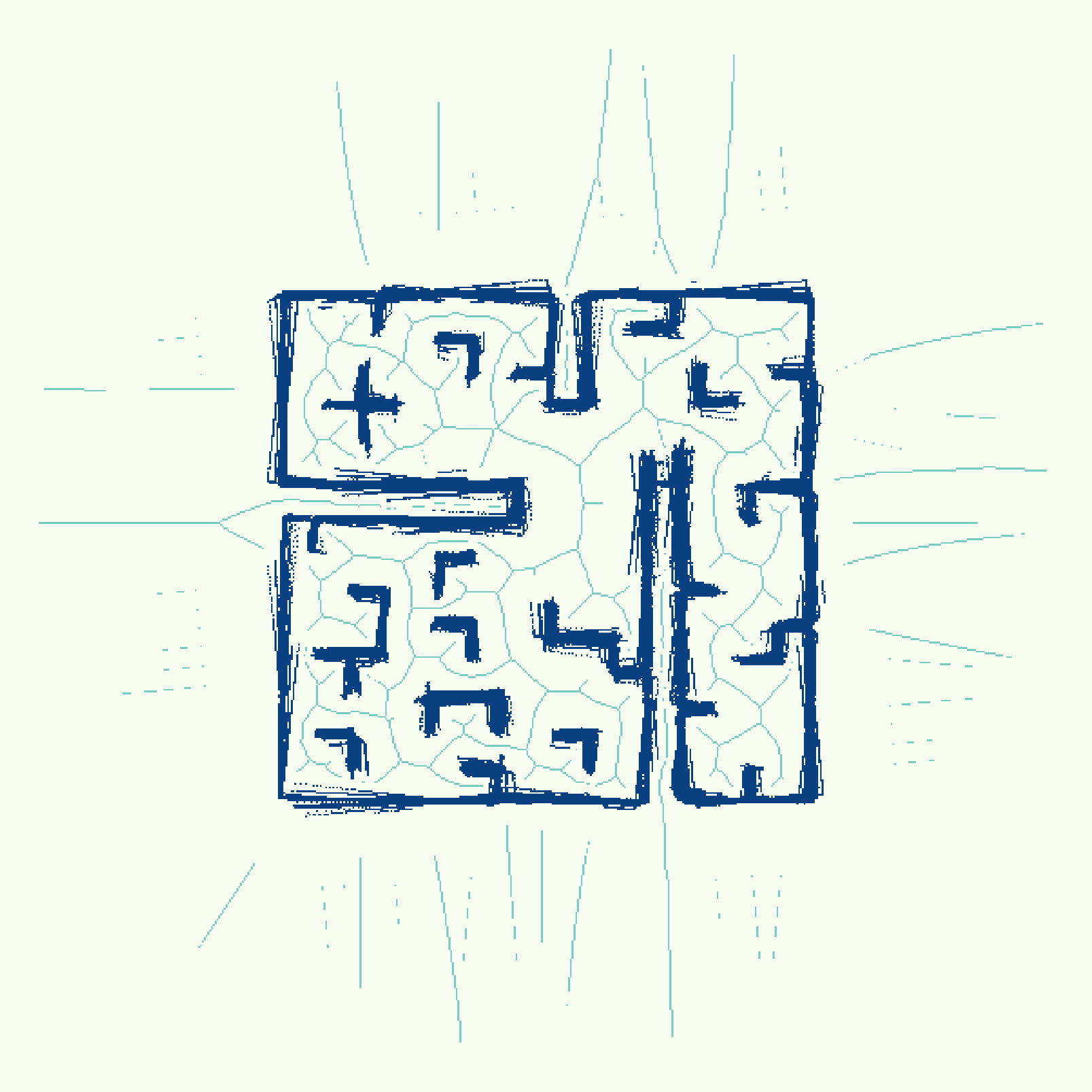}
	\includegraphics[width=\blaSize\linewidth]{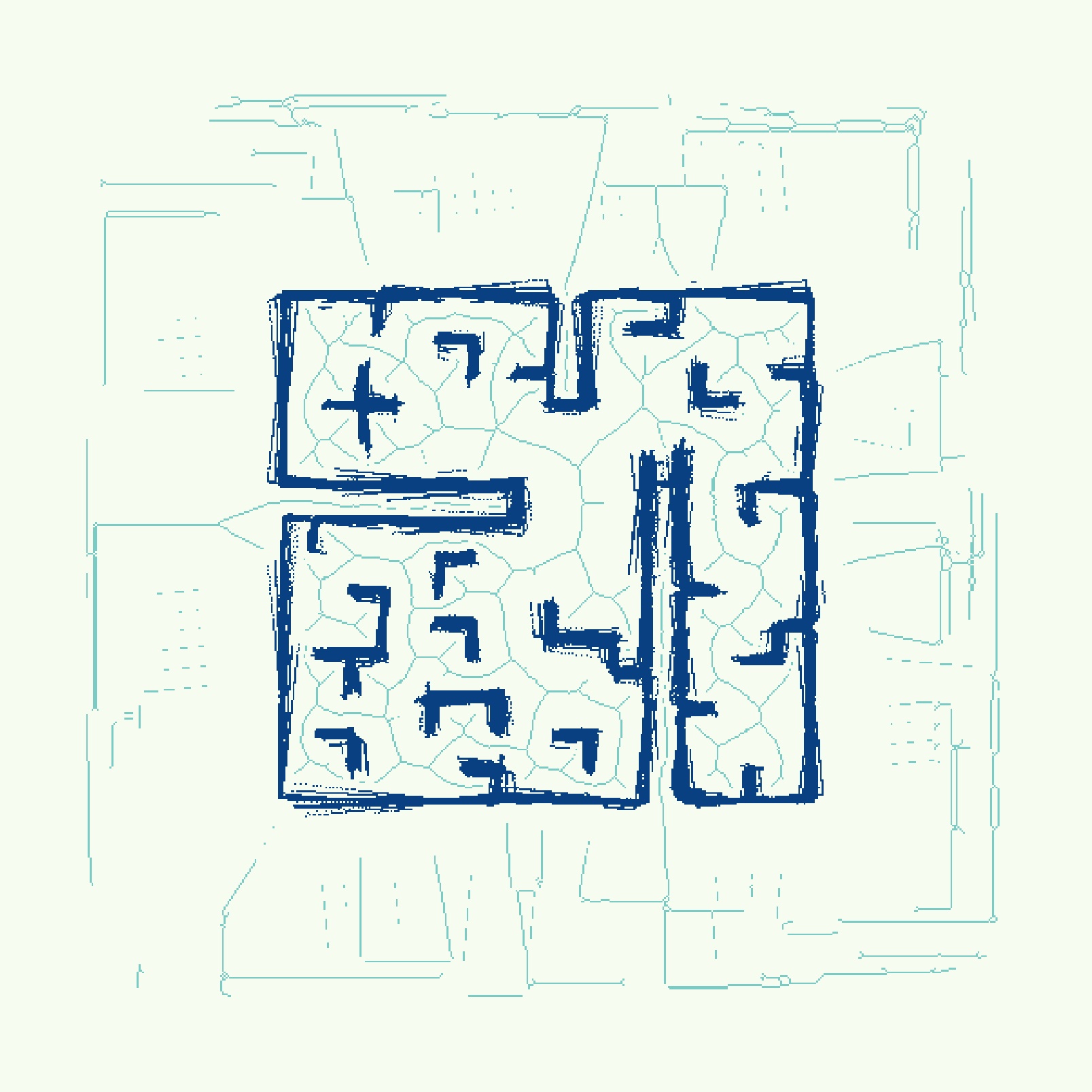}\\ \vspace{1mm}
	\includegraphics[width=\blaSize\linewidth]{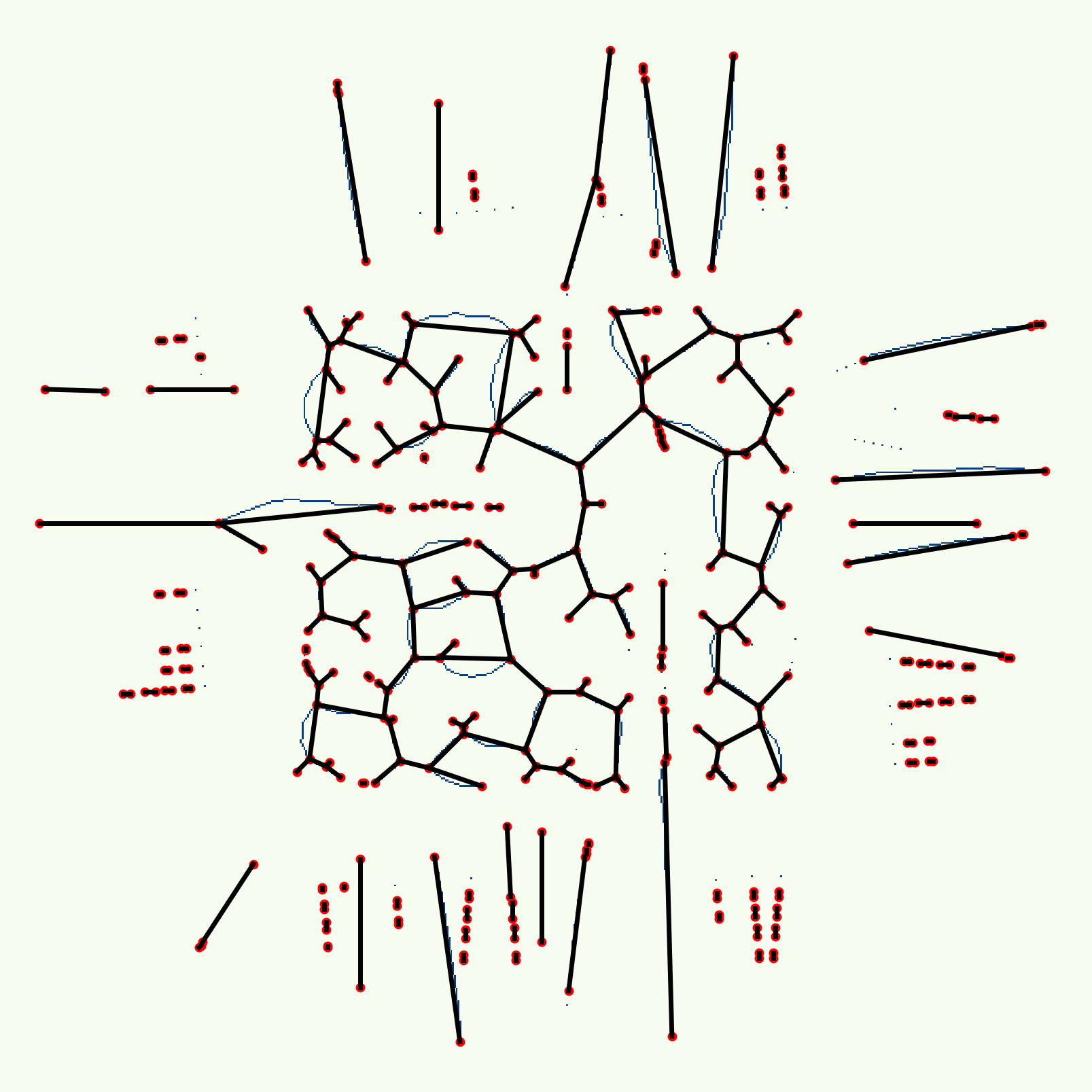}
	\includegraphics[width=\blaSize\linewidth]{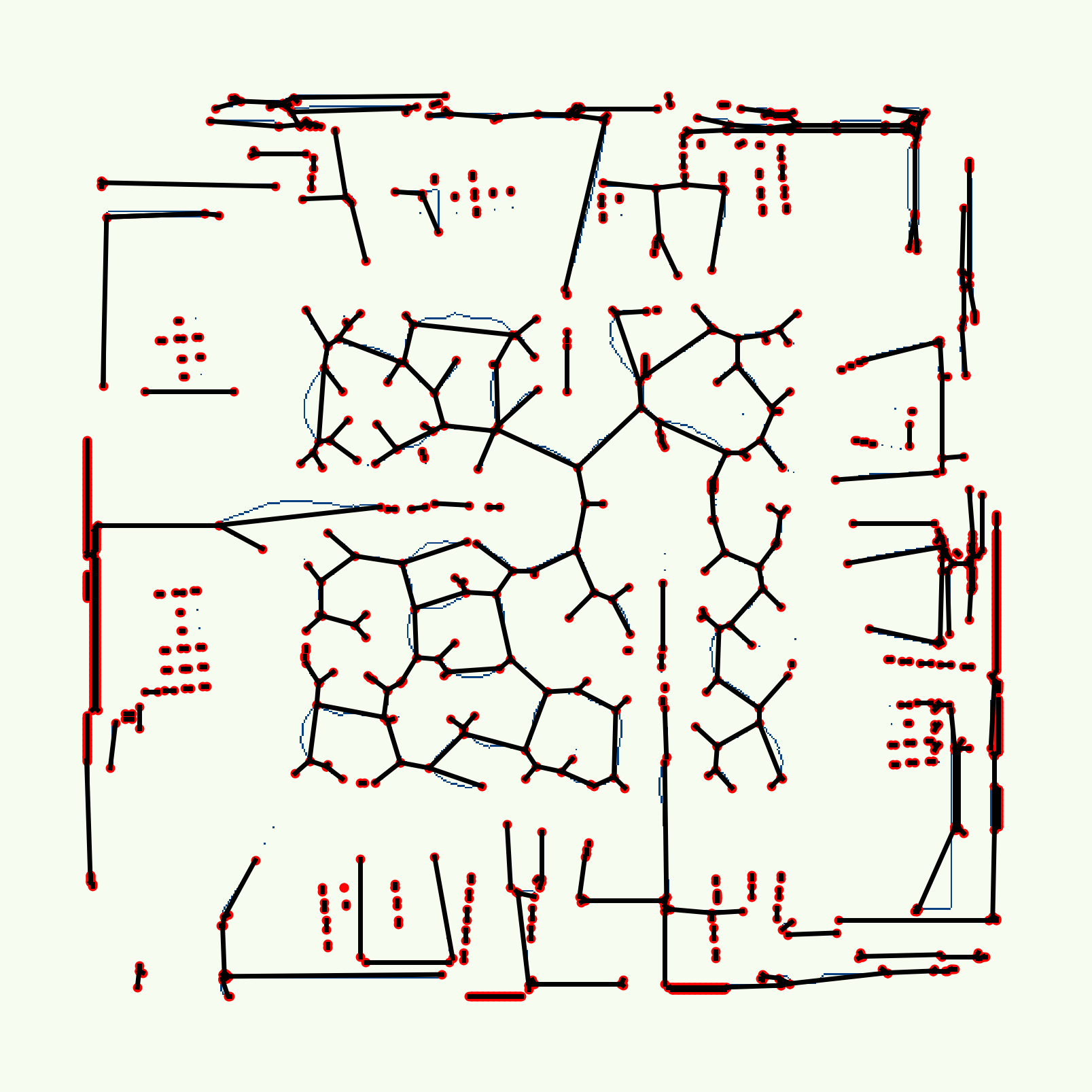}
	\caption{Comparing the full map results with incremental algorithm results. The first column are the results with final occupancy input and second column is from our incremental framework. The three rows are for distance map, skeleton and topological map, respectively.}
	\label{fig:framesincre}
\end{figure}

\newcommand\blubSize{0.18}
\begin{figure*}[]
	\centering
	\includegraphics[width=\blubSize\linewidth]{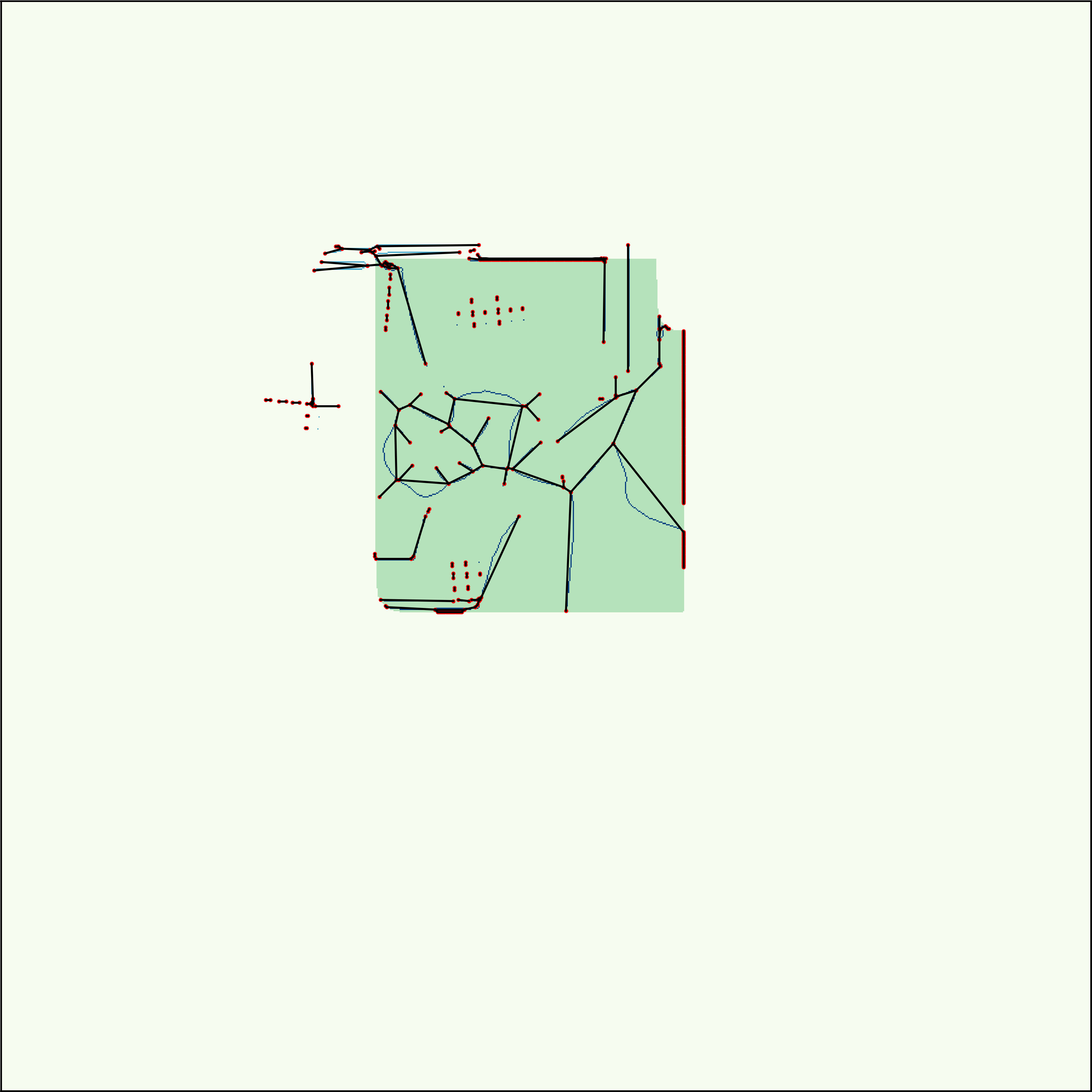}
	\includegraphics[width=\blubSize\linewidth]{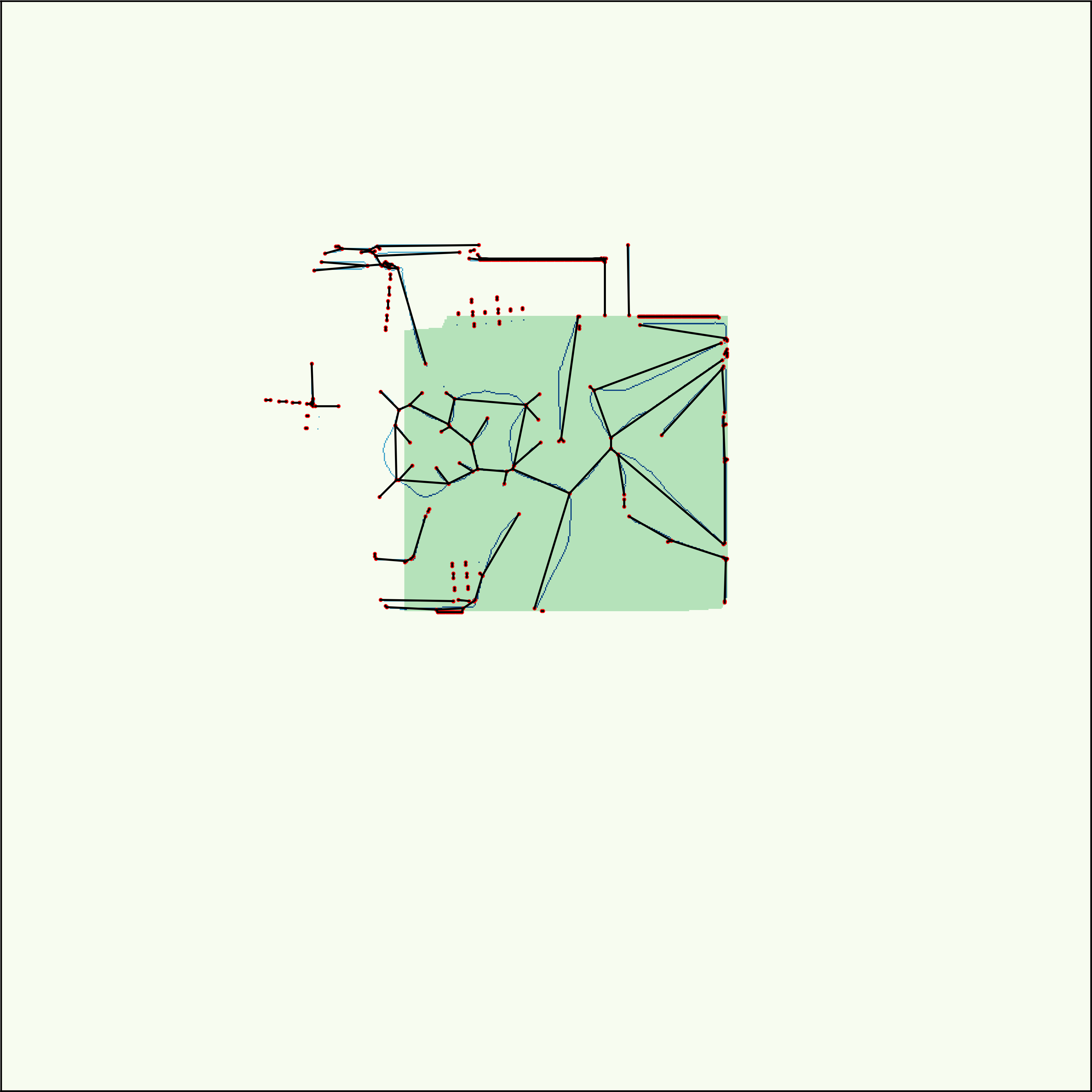}
	\includegraphics[width=\blubSize\linewidth]{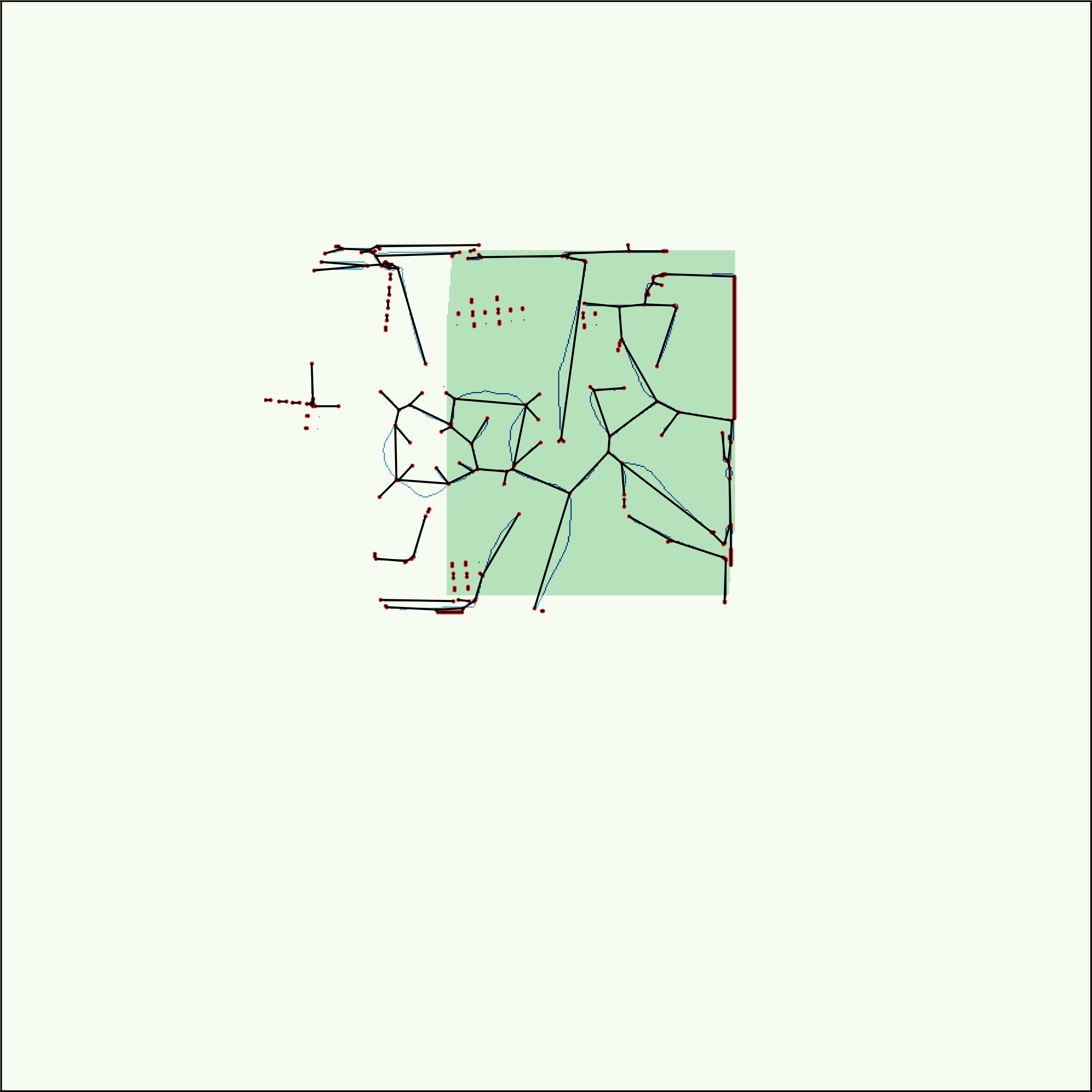}
	\includegraphics[width=\blubSize\linewidth]{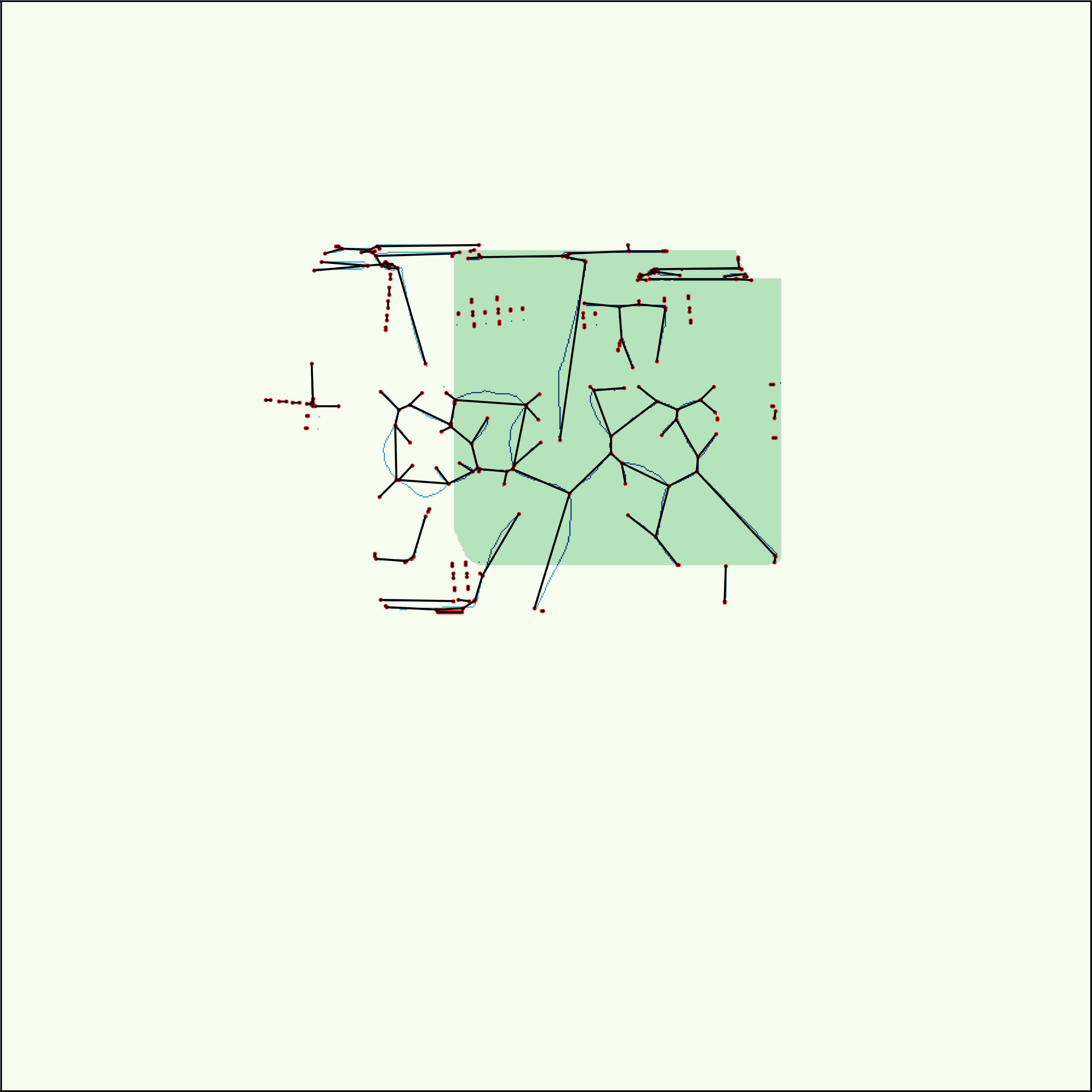}
	\includegraphics[width=\blubSize\linewidth]{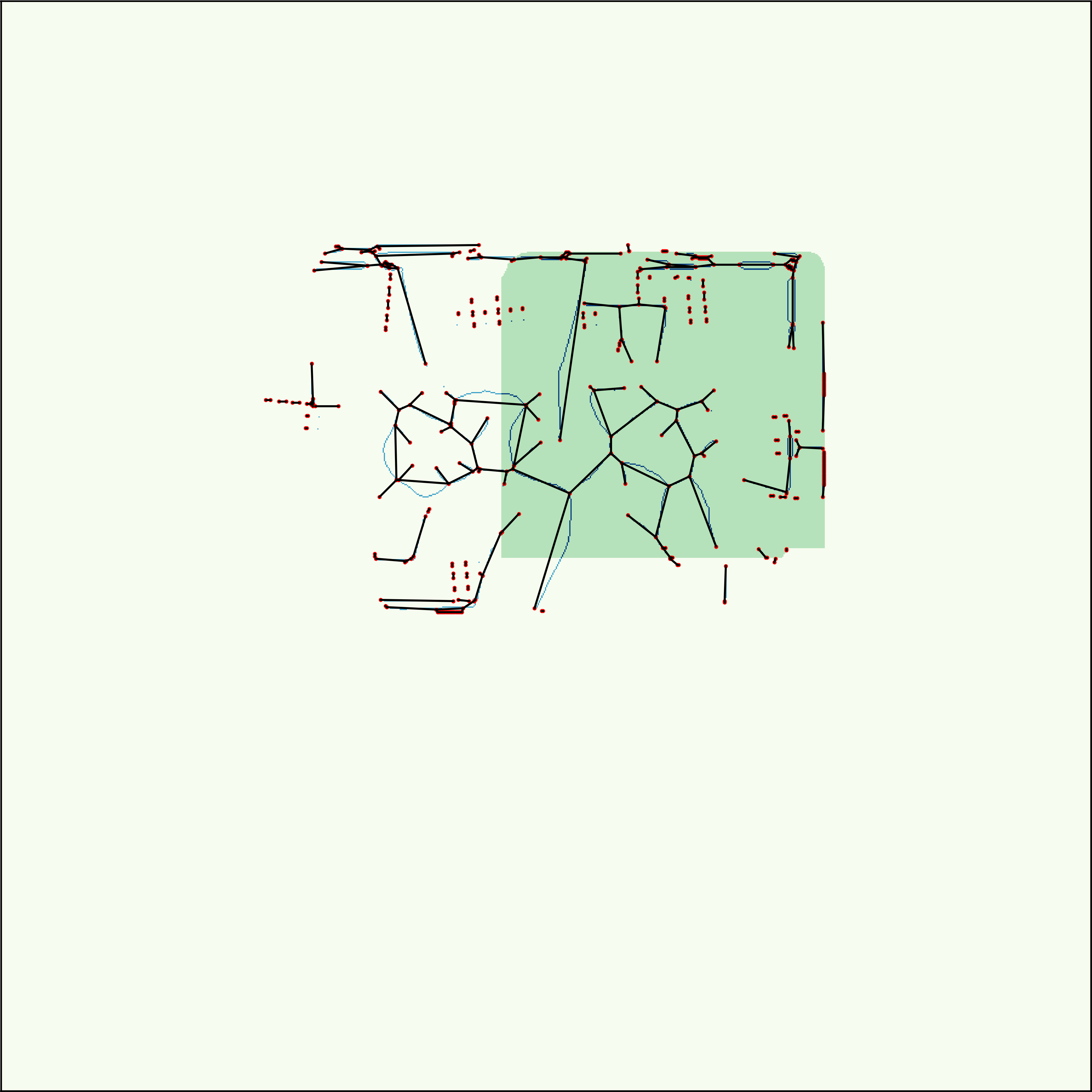}\\\vspace{1mm}
	\includegraphics[width=\blubSize\linewidth]{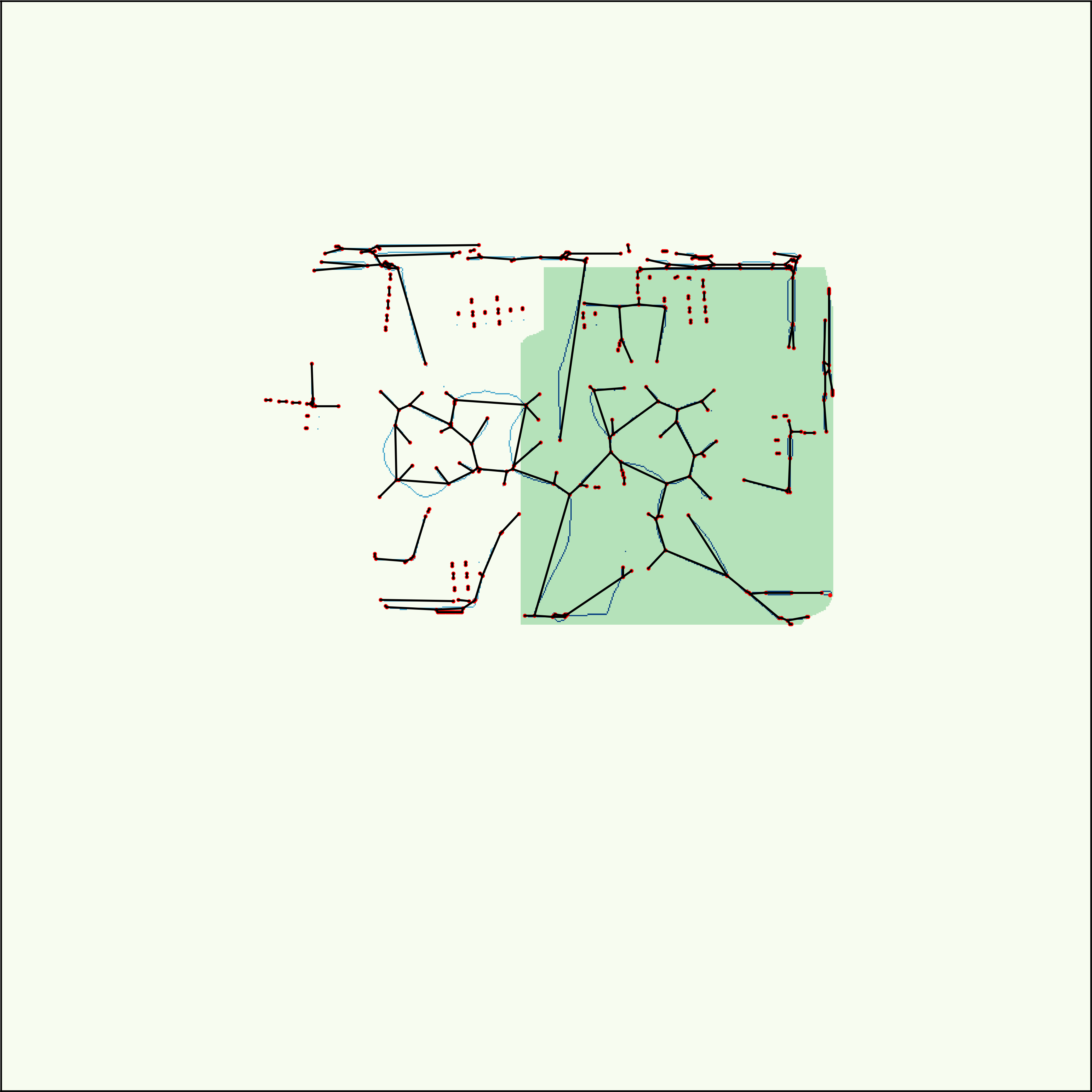}
	\includegraphics[width=\blubSize\linewidth]{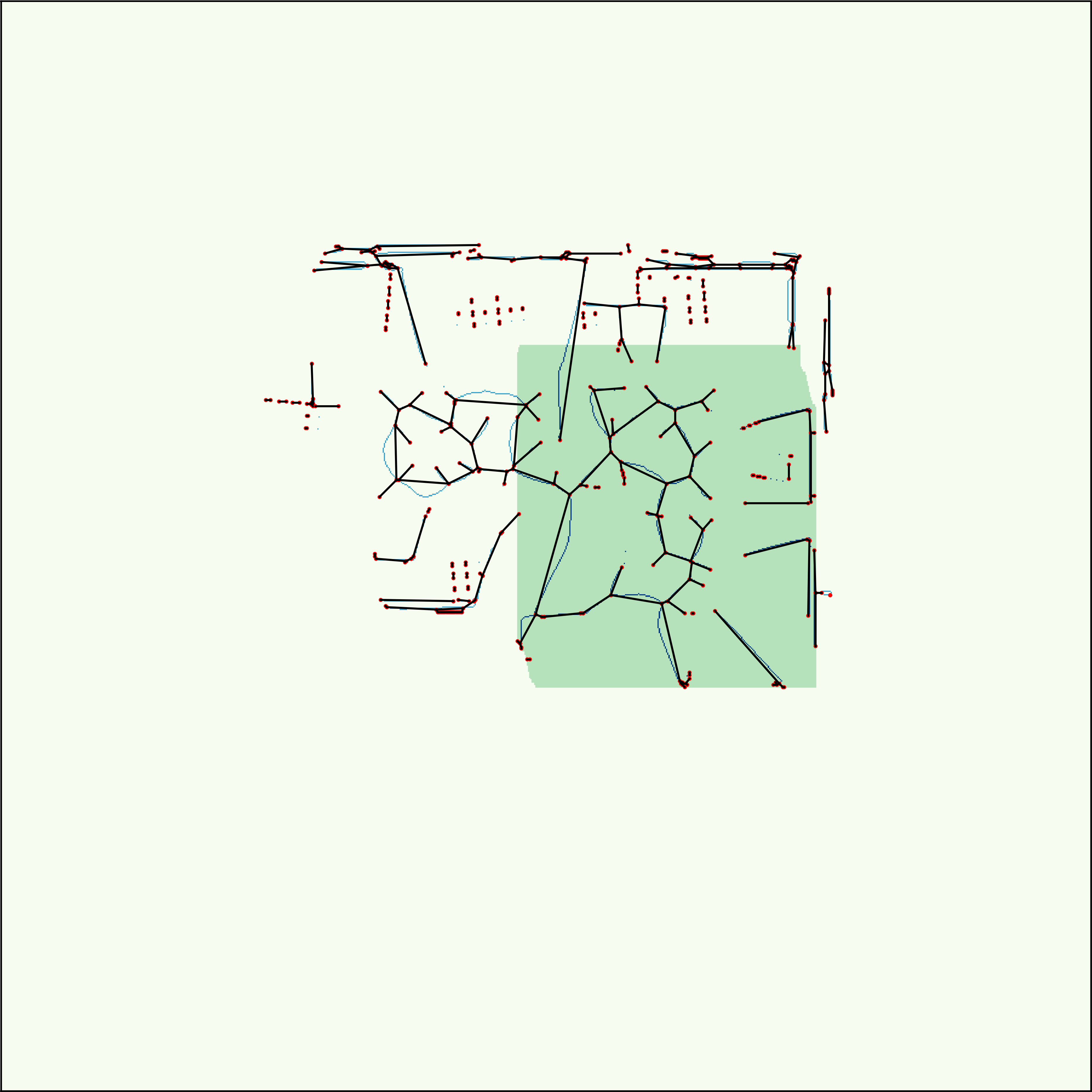}
	\includegraphics[width=\blubSize\linewidth]{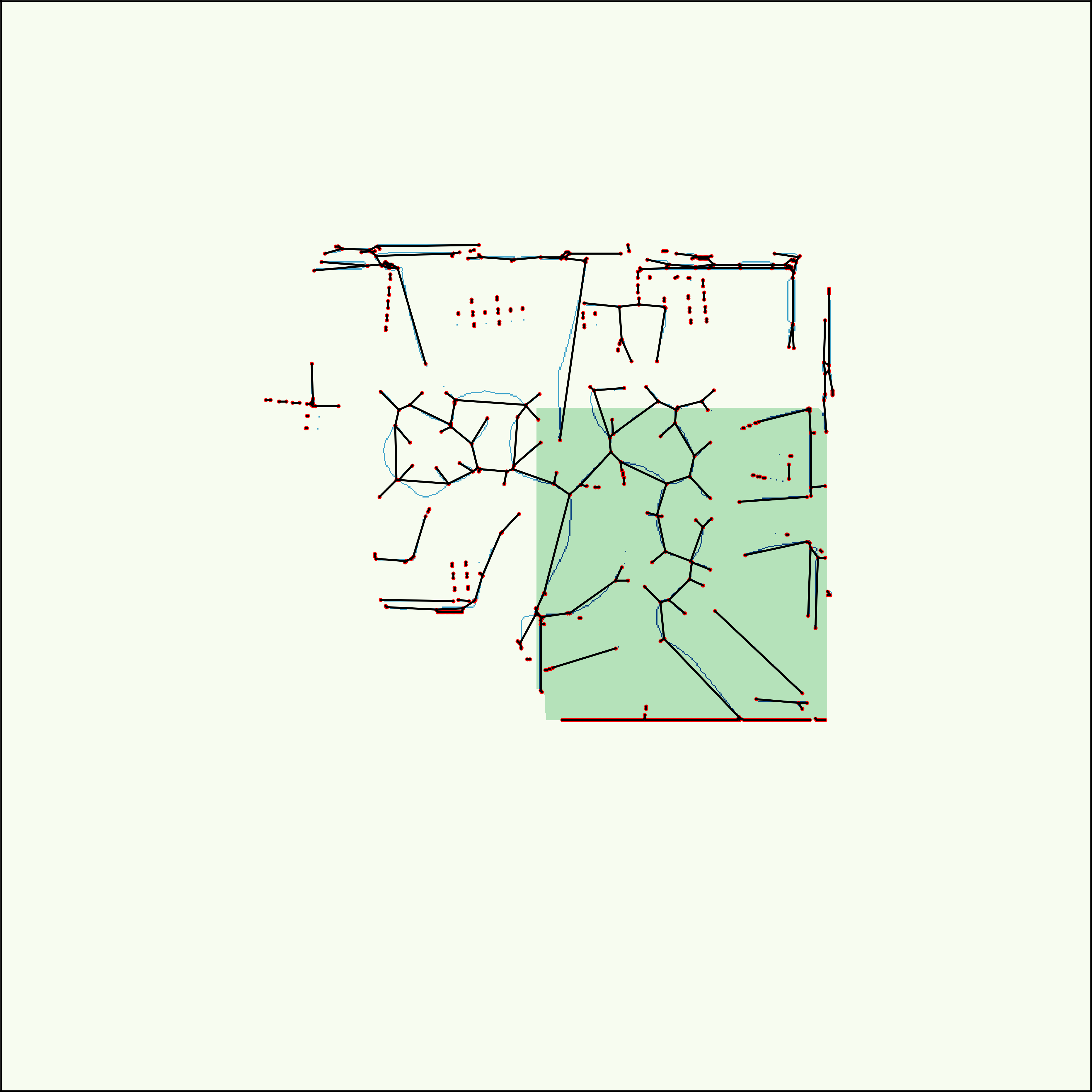}
	\includegraphics[width=\blubSize\linewidth]{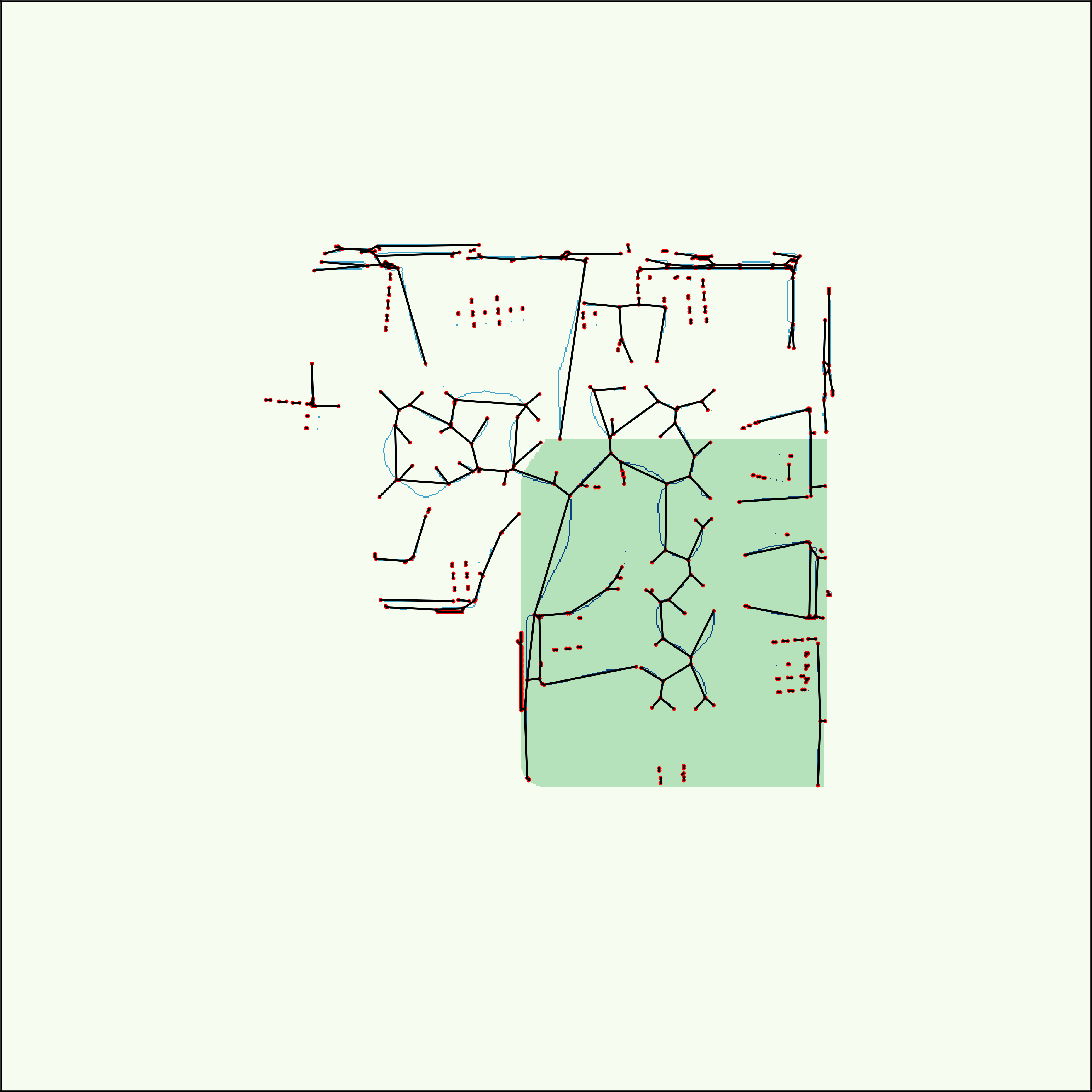}
	\includegraphics[width=\blubSize\linewidth]{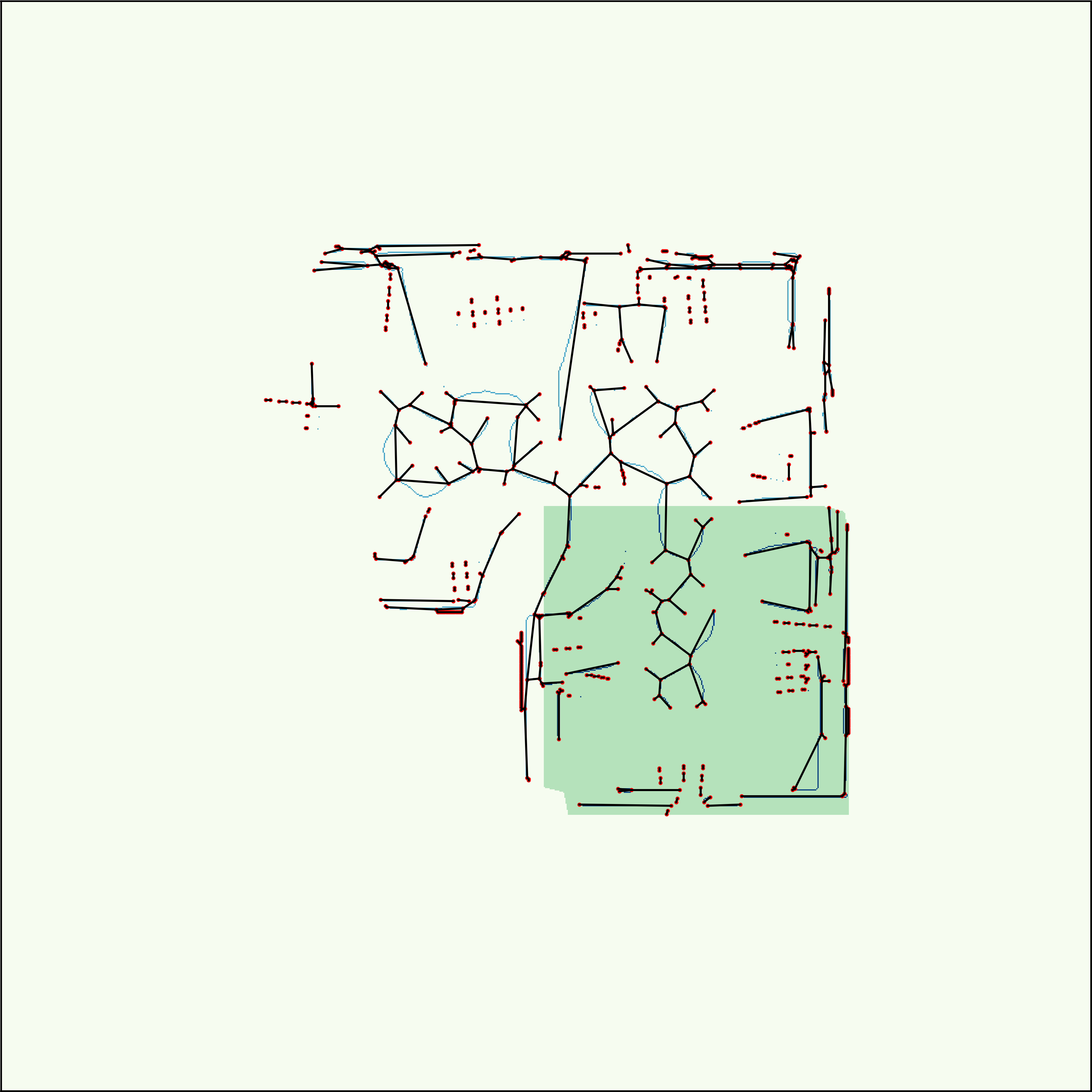}
	\caption{Ten sequential updates of the incrementally build topological map of Fig. \ref{fig:framesincre}. The deep green region is the masked local region described in Section \ref{sec:33}. The selected maps are with frame ID 400 to 1120 with a step of 80. See the accompanying video for all skeleton and topological maps for this experiment.}
	\label{fig:middleFrames}
\end{figure*}

We realize our framework with python. During the implementation, Scikit-image\cite{van2014scikit}, OpenCV and ROS have been utilized. We provide the source on Github \footnote{\url{https://github.com/STAR-Center/IncrementalTopo}}.

It is important to mention some detail of our implementation. The distance map has been generated with a Gaussian kernel, and its pixel hold quite small values. Thus, before passing the Laplacian filter, we scale it with 255. To obtain the binary grid map, we utilized a threshold, 10, and set bigger values True, and False otherwise. To extract the skeleton, we apply the $thin$ function from Scikit-image to the binary image. 

The map utilized in the experiment of Section \ref{sec:exp2} is generated with a simulated Hokuyo laser scanner in STDR simulator (simple two dimensional robot simulator) under ROS. In STDR, accurate odometry can be extracted and its laser scan data has a Gaussian noise with mean 0.5 and standard deviation 0.05. The ROS-bag we recorded contains 3061 frames in total. For our incremental algorithm, from the beginning, we update the distance map every frame, the skeleton every 20 frames and the topological map every 80 frames. 3040 frames are used in total.

Our experiment is run on a PC with Intel Core $i7-7700$ $3.6GHz$ with Ubuntu 16.04.

\subsection{Experiment on Building Topological Maps}
\label{sec:exp1}

We consider it necessary to compare the quality of the global approach from Section \ref{sec:2} with a common Voronoi diagram based method. 

We apply both our code and the Voronoi diagram based method implemented in \cite{schwertfeger2016map} to three maps. 
The topological maps are annotated with path information (from Voronoi diagram or skeleton), because it keeps the vital information that can describe how the paths look.

From Fig. \ref{fig:vori_vs_heatmap} we can find, that our implementation achieved a similar result as the Voronoi based method. The structure is the same and, more importantly, the positions of the skeleton pixels are very close to the common Voronoi diagram based method. As in Fig. \ref{fig:vori_vs_heatmap}, for each vertex of topological graph in the right column map, we find its closest vertex in the left column map to compute its error, which is the euclidean distance. Note, that we consider vertexes where the closest distance is larger than 20 pixel as outliers. The average distance of non-outlier vertexes, number of outlier and total vertexes, percentage of vertexes with distance smaller or equal than $1$ can be found in Table \ref{tab:avg_time}. It shows a very low average distance. For the intel map, the average distance is $2.17$, larger than the other two maps, but $72.4\%$ of the vertexes are within one pixel. For the office map, the percentage is only $49.5\%$. However, it's average is only $1.35$ pixel. $a_scan$ has a very low error of $0.88$ with $84.9\%$ of vertexes within one pixel.

This shows that the distance map based approach is a good basis for our incremental framework.

\setlength{\belowcaptionskip}{0pt}

\subsection{Experiments with the Incremental Framework}
\label{sec:exp2}
\begin{figure}[!]
	\centering
	\includegraphics[width=1.\linewidth]{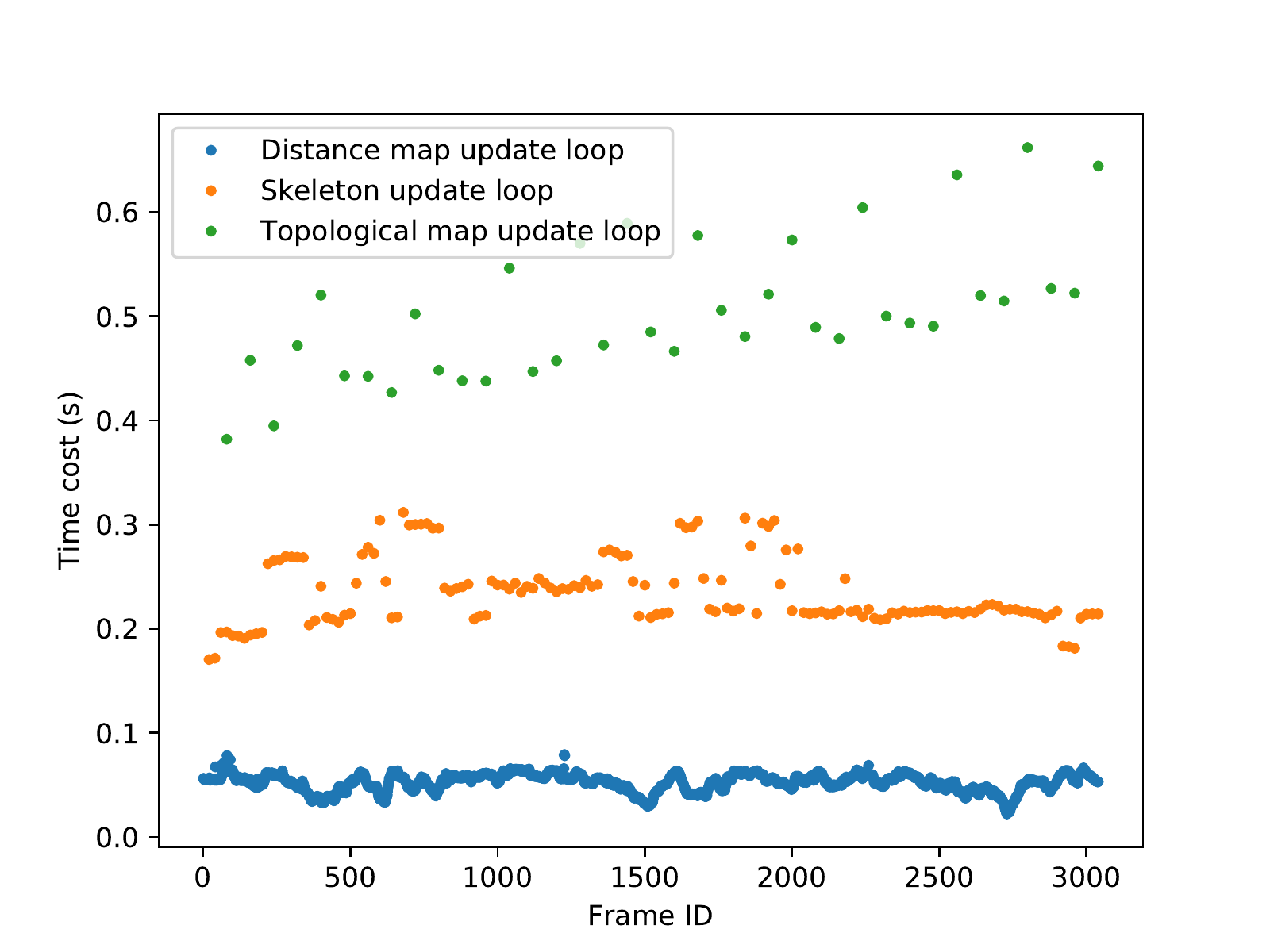}
	\caption{Time cost of each step for our incremental algorithm. }
	\label{fig:time_table}
\end{figure}

Computing time evaluations are essential for real-time algorithms in robotics. We record the update cost (without drawing the map) in our framework and illustrate this point in Fig. \ref{fig:time_table}. As described in Section \ref{sec:setting}, the distance map, skeleton and topological map are updated every 1, 20 and 80 frames, respectively. For those three update loops,  the later loop is based on its former loop. However, the former loop does not depend on the later one, which make it not necessary to sequentially update every time. What's more, they could even be in different threads for the implementation. Thus, we consider the time to update the distance map should be the most important. Generally, the range sensor provides $10Hz$ scans. From Fig. \ref{fig:time_table}, the distance map update takes no more than $0.1s$ per frame, which meets our needs.

For the incremental topological map generation, our goal in this experiment is to compare the quality with the algorithm in Section \ref{sec:2}.

Fig. \ref{fig:framesincre} shows the result. We can find that the inside parts of the room are almost the same. The major difference between those two maps are in the outside areas of the room. Those outside areas are not important for the applications using the topological maps later, because they are outside the scope of the mapped area. 

Actually, as in \cite{schwertfeger2016map}, some algorithms like alpha shape can be utilized to delete the edges outside the room. The effect of the additional outer edges could be cleared in this way without problem.

We also compute the vertex distance (error) between the non-incremental and incremental graphs in Fig. \ref{fig:framesincre}, as we did in Section \ref{sec:exp1}. Only considering the inside room region that is from pixel 300 to 600 on both axises, the average distance of the vertexes is $2.28$ pixel with $5$ outliers from $251$ vertexes and $63.0\%$ vertexes within one pixel which is invisibly small for incremental topology graph and thus is better than match and merge based method.

We also show ten images of sequential updates of the topological map in the middle of incremental map building in Fig. \ref{fig:middleFrames}, to demonstrate the incremental update of the topological map. In these figures, we can see that the newly updated region does not affect the non-touched region. One advantage we can find is, that the updating region is with the current global distance map, which is the most acceptable topology graph at this moment.


\section{Conclusions}
\label{sec:conclusions}
In this work, we presented the distance map based method to generate topological maps and, on top of it, further proposed a framework that can work well with on-going sensor scanning to build the topological map incrementally. Our experiments show the performance of the presented approach. We found, that our extracted maps are similar to those of the commonly used Vorinoi Diagram based method. We embed the distance map based approach into our incremental framework and found that the update speed is fast and the resulting topological maps are almost the same when compared with the non-incremental version. We thus showed that it is possible to incrementally build and update a topological maps while incorporating sensor data in real time.

\IEEEtriggeratref{3}

\bibliographystyle{IEEEtran}
\bibliography{references.bib}



\end{document}